\documentclass{article}

\usepackage{microtype}
\usepackage{graphicx}
\usepackage{subfigure}
\usepackage{booktabs} % for professional tables

\usepackage{hyperref}

\usepackage[accepted]{icml2023}

\usepackage{amsmath}
\usepackage{amssymb}
\usepackage{mathtools}
\usepackage{amsthm}
\usepackage{bm, float}

\usepackage[capitalize,noabbrev]{cleveref}

\theoremstyle{plain}

\theoremstyle{definition}

\theoremstyle{remark}

\usepackage[textsize=tiny]{todonotes}

\icmltitlerunning{Automatic Intrinsic Reward Shaping for Exploration in Deep Reinforcement Learning}

\begin{document}

\twocolumn[
\icmltitle{Automatic Intrinsic Reward Shaping for Exploration in Deep Reinforcement Learning}

% It is OKAY to include author information, even for blind
% submissions: the style file will automatically remove it for you
% unless you've provided the [accepted] option to the icml2023
% package.

% List of affiliations: The first argument should be a (short)
% identifier you will use later to specify author affiliations
% Academic affiliations should list Department, University, City, Region, Country
% Industry affiliations should list Company, City, Region, Country

% You can specify symbols, otherwise they are numbered in order.
% Ideally, you should not use this facility. Affiliations will be numbered
% in order of appearance and this is the preferred way.
\icmlsetsymbol{equal}{*}

\begin{icmlauthorlist}
\icmlauthor{Mingqi Yuan}{hkpu}
\icmlauthor{Bo Li}{hkpu}
\icmlauthor{Xin Jin}{eias}
\icmlauthor{Wenjun Zeng}{eias}
%\icmlauthor{Firstname1 Lastname1}{equal,yyy}
%\icmlauthor{Firstname2 Lastname2}{equal,yyy,comp}
%\icmlauthor{Firstname3 Lastname3}{comp}
%\icmlauthor{Firstname4 Lastname4}{sch}
%\icmlauthor{Firstname5 Lastname5}{yyy}
%\icmlauthor{Firstname6 Lastname6}{sch,yyy,comp}
%\icmlauthor{Firstname7 Lastname7}{comp}
%\icmlauthor{}{sch}
%\icmlauthor{Firstname8 Lastname8}{sch}
%\icmlauthor{Firstname8 Lastname8}{yyy,comp}
%\icmlauthor{}{sch}
%\icmlauthor{}{sch}
\end{icmlauthorlist}

\icmlaffiliation{hkpu}{Department of Computing, The Hong Kong Polytechnic University, Hong Kong, China}
\icmlaffiliation{eias}{Eastern Institute for Advanced Study, Zhejiang, China}
\icmlcorrespondingauthor{Xin Jin}{jinxin@eias.ac.cn}
%\icmlaffiliation{yyy}{Department of XXX, University of YYY, Location, Country}
%\icmlaffiliation{comp}{Company Name, Location, Country}
%\icmlaffiliation{sch}{School of ZZZ, Institute of WWW, Location, Country}
%
%\icmlcorrespondingauthor{Firstname1 Lastname1}{first1.last1@xxx.edu}
%\icmlcorrespondingauthor{Firstname2 Lastname2}{first2.last2@www.uk}

% You may provide any keywords that you
% find helpful for describing your paper; these are used to populate
% the "keywords" metadata in the PDF but will not be shown in the document
\icmlkeywords{Machine Learning, ICML}

\vskip 0.3in
]

% this must go after the closing bracket ] following \twocolumn[ ...

% This command actually creates the footnote in the first column
% listing the affiliations and the copyright notice.
% The command takes one argument, which is text to display at the start of the footnote.
% The \icmlEqualContribution command is standard text for equal contribution.
% Remove it (just {}) if you do not need this facility.

%\printAffiliationsAndNotice{}  % leave blank if no need to mention equal contribution
\printAffiliationsAndNotice{} % otherwise use the standard text.

\begin{abstract}
We present AIRS: \textbf{A}utomatic \textbf{I}ntrinsic \textbf{R}eward \textbf{S}haping that intelligently and adaptively provides high-quality intrinsic rewards to enhance exploration in reinforcement learning (RL). More specifically, AIRS selects shaping function from a predefined set based on the estimated task return in real-time, providing reliable exploration incentives and alleviating the biased objective problem. Moreover, we develop an intrinsic reward toolkit \footnote{\url{https://github.com/RLE-Foundation/rllte}} to provide efficient and reliable implementations of diverse intrinsic reward approaches. We test AIRS on various tasks of MiniGrid, Procgen, and DeepMind Control Suite. Extensive simulation demonstrates that AIRS can outperform the benchmarking schemes and achieve superior performance with simple architecture.
\end{abstract}

\section{Introduction}
Striking an appropriate balance between exploration and exploitation remains a long-standing problem in reinforcement learning (RL) \cite{sutton2018reinforcement}. Sufficient exploration can prevent the RL agent from prematurely falling local optima after finite iterations, contributing to learning better policies \cite{franccois2018introduction}. To address this problem, classical exploration strategies such as $\epsilon$-greedy or Boltzmann exploration will randomly choose all the possible actions with a non-zero probability \cite{mnih2015human}. But these approaches are inefficient when handling complex environments with high-dimensional observations. Moreover, extrinsic rewards are consistently found to be sparse or even absent in many real-world scenarios, and they may completely fail to learn.

Recent approaches have proposed to leverage intrinsic rewards to encourage exploration \cite{oh2015action, houthooft2016vime, bellemare2016unifying, pathak2017curiosity, haber2018learning, ostrovski2017count}. For instance, \cite{bellemare2016unifying} leverage a density model to approximate the state visitation frequency and define the intrinsic reward as inversely proportional to the pseudo-count. As a result, the agent is encouraged to visit the infrequently-seen states, increasing the probability of encountering states with higher task rewards. In contrast, curiosity-driven exploration aims to learn the dynamics of the environment and utilizes the prediction error as the intrinsic reward \cite{stadie2015incentivizing,yu2020intrinsic,burda2018exploration}. For example, \cite{pathak2017curiosity} use an inverse-forward dynamics model to learn the representation of state space, which only encodes the part that affects the decision-making and ignores environment noise and other irrelevant interference. After that, the intrinsic reward is defined as the prediction error of the encoded next-state based on the current state-action pair.

\begin{figure}[t]
	\vskip 0.2in
	\begin{center}
		\centerline{\includegraphics[width=0.95\linewidth]{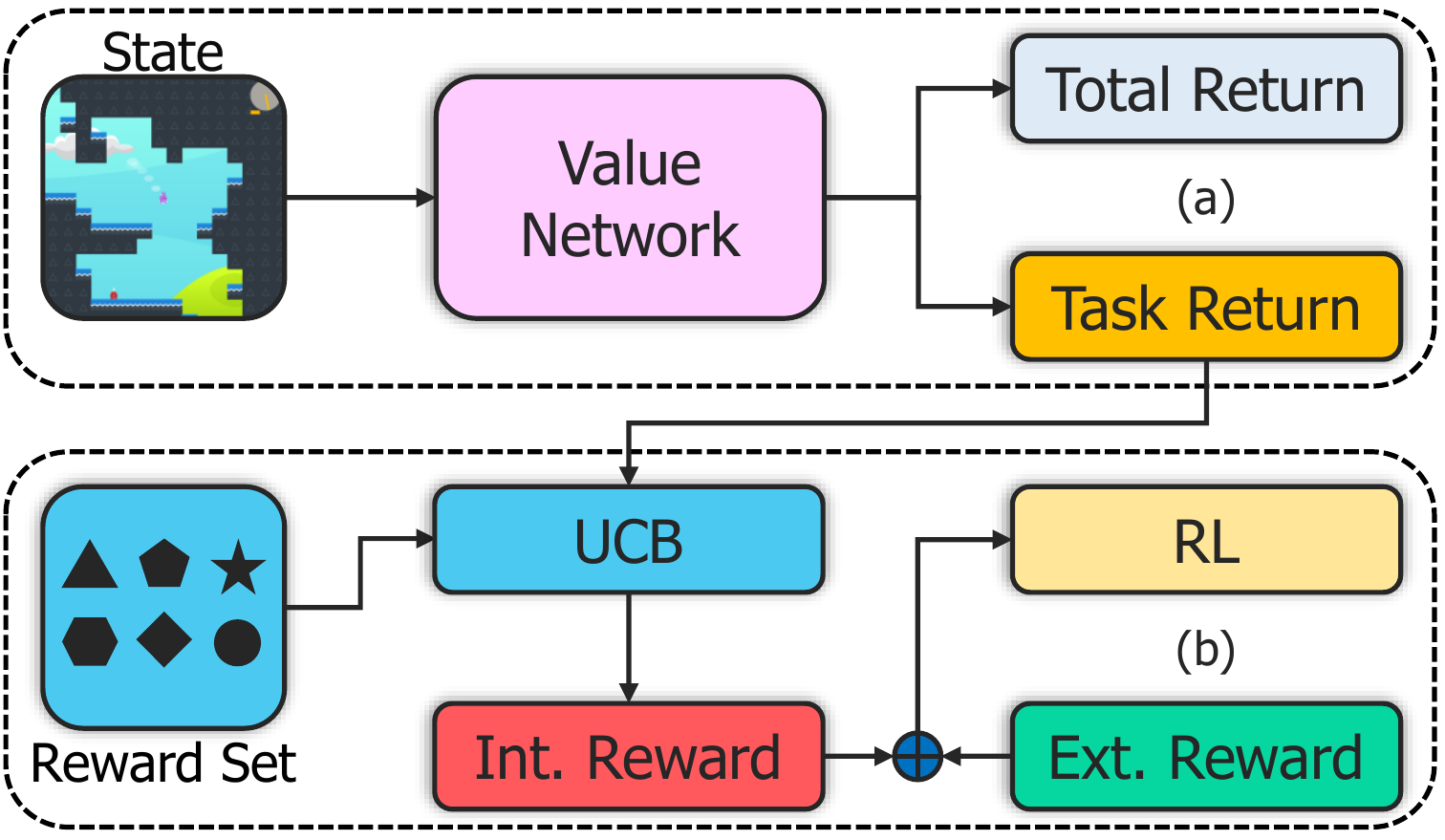}}
		\caption{(a) AIRS employs two branches in the value network to disentangle the estimation of total return (evaluated by intrinsic and extrinsic reward function) and task return (only extrinsic reward function). (b) AIRS formulates the intrinsic reward selection as a multi-armed bandit problem and uses the upper confidence bound (UCB) to make decisions based on the estimated task return. \textbf{Ext.}: Extrinsic. \textbf{Int.}: Intrinsic. $\bigoplus$: Weighted summation operation.}
	\label{fig:guide}
\end{center}
\vskip -0.2in
\end{figure}

Despite the excellent performance of intrinsic rewards, they cannot guarantee the invariance of the optimal policy, and excessive exploration may lead to learning collapse \cite{ng1999policy}. Since the joint objective composed of extrinsic and intrinsic rewards is biased, maximizing it only sometimes yields the optimal policy for the extrinsic reward alone. To alleviate this problem, \cite{chen2022redeeming} proposed an extrinsic-intrinsic policy optimization (EIPO) approach, which automatically tunes the importance of the intrinsic reward via a principled constrained policy optimization procedure. As a result, EIPO can restrain the intrinsic reward when exploration is unnecessary and increase it when exploration is required. However, EIPO is too sophisticated to implement and cannot provide increments for arbitrary RL algorithms. Moreover, many experiments reported in existing work \cite{burda2018large,burda2018large,raileanu2020ride,badia2020never,chen2022redeeming} demonstrate that the performance of the same intrinsic rewards varies significantly in different tasks and learning stages. Selecting the best intrinsic reward for a specific task is always challenging. With this in mind, \cite{zheng2018learning} proposed a stochastic gradient-based method to learn parametric intrinsic rewards, which improved the performance of policy gradient-based algorithms.

In this paper, we solve the aforementioned problems using a novel framework entitled \textbf{A}utomatic \textbf{I}ntrinsic \textbf{R}eward \textbf{S}haping (AIRS). Our main contributions can be summarized as follows:
\begin{itemize}
\item Firstly, we formulate the intrinsic reward selection as a multi-armed bandit problem, in which different intrinsic reward functions are regarded as arms. AIRS can automatically select the best shaping function at different learning stages based on the estimated task return, providing reliable exploration incentives and alleviating the biased objective problem.
\item Secondly, we develop a toolkit that provides high-quality implementations of various intrinsic reward modules based on PyTorch. These modules can be deployed in arbitrary RL algorithms in a plug-and-play manner, providing efficient and robust exploration increments.
\item Finally, we test AIRS on MiniGrid, Procgen (\textit{sixteen games with procedurally-generated environments}), and DeepMind Control Suite. Extensive simulation results demonstrate that AIRS can achieve superior performance and generalization ability and outperform the benchmarking schemes.
\end{itemize}

\section{Related Work}
\subsection{Count-Based Exploration}
Count-based exploration provides intrinsic rewards by measuring the novelty of states, which are usually defined to be inversely proportional to the state visit counts \cite{bellemare2016unifying, ostrovski2017count, tang2017exploration, machado2020count, jo2022leco}. \cite{strehl2008analysis} proposed to use the state visit counts as exploration bonuses in the tabular setting and provided a theoretical explanation for its effectiveness. \cite{bellemare2016unifying} designed a context-tree switching density model to perform state pseudo-count, and \cite{ostrovski2017count} considered the environments with high-dimensional observations and used a PixelCNN as a state density estimator. \cite{martin2017count} proposed to perform pseudo-count by leveraging the learned feature space of value function approximation, which can evaluate the uncertainty associated with any state. \cite{burda2018exploration} proposed a random-network-distillation (RND) method composed of a predictor network and a target network, using the prediction error to reward novel states. 

\subsection{Curiosity-Driven Exploration}
Curiosity-driven exploration encourages the agent to increase its knowledge (e.g., dynamics) about the task environment \cite{stadie2015incentivizing, pathak2017curiosity, yu2020intrinsic}. The most well-known work is the intrinsic-curiosity-module (ICM) proposed by \cite{pathak2017curiosity}. However, \cite{yu2020intrinsic} suggested that the discriminative model may suffer from the compounding error and redesigned the architecture of ICM using a variational auto-encoder \cite{kingma2013auto}, which circumvents the encoding of state space and can be trained using a one-life demonstration. In particular, \cite{burda2018large} attempted to solve visual tasks (e.g., Atari games) using only curiosity-based intrinsic reward, and the agent could still achieve remarkable performance. But the curiosity-driven methods are consistently found to be futile when handling the noisy-TV dilemma \cite{savinov2018episodic}. To address the problem, \cite{raileanu2020ride} proposed a rewarding-impact-driven-exploration (RIDE) method that uses the difference between two consecutive encoded states as the intrinsic reward and encourages the agent to choose actions that result in significant state changes. In this paper, the RIDE is selected as the candidate for the intrinsic reward set. It can provide aggressive exploration incentives and oblige the agent to adapt to the environment quickly, improving the generalization ability of the agent and facilitating solving tasks with procedurally-generated environments.

\begin{figure*}[t]
	\vskip 0.2in
	\begin{center}
		\centerline{\includegraphics[width=0.85\linewidth]{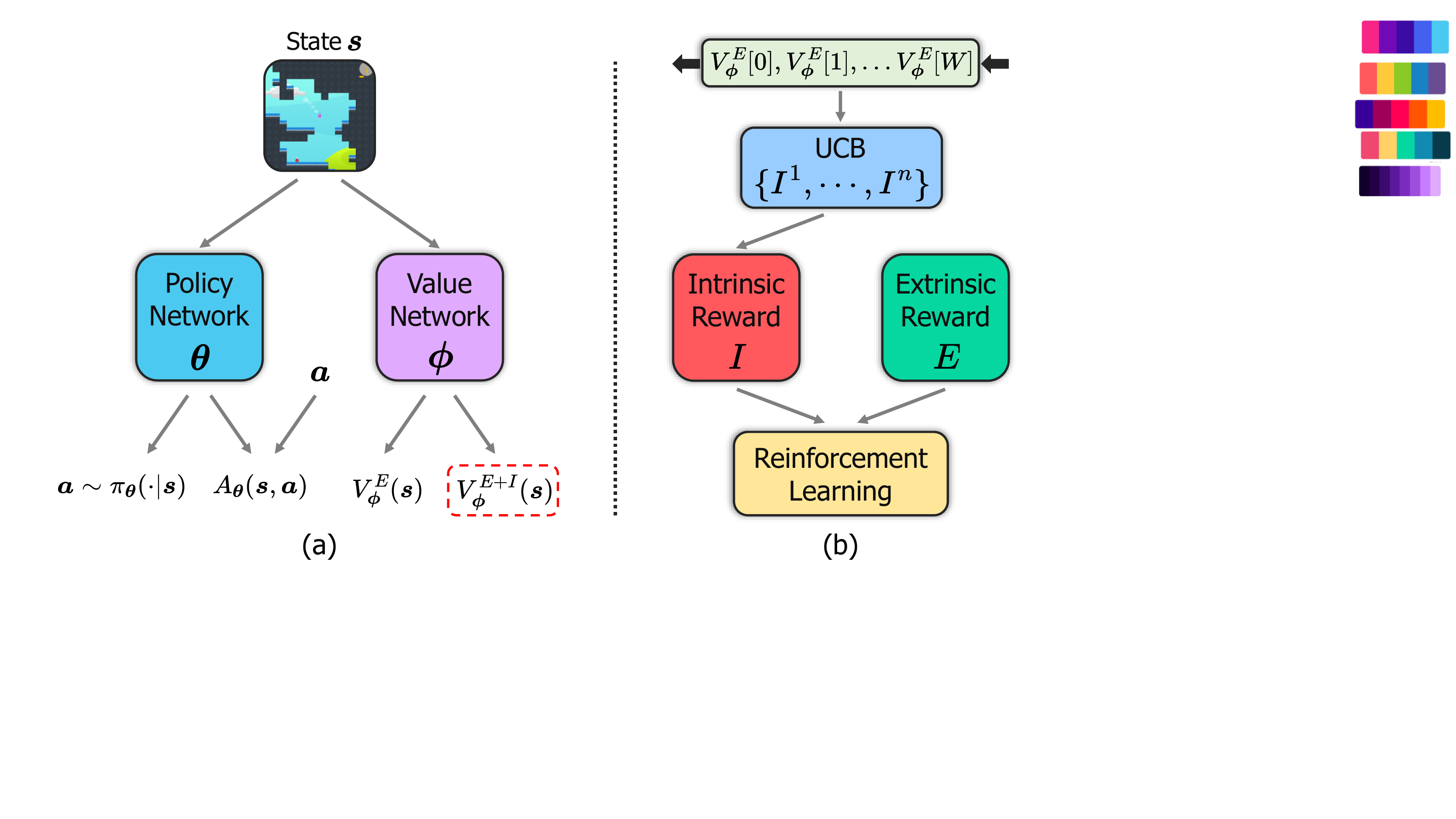}}
		\caption{The overview of AIRS. (a) The architecture of DAAC. Here, we use two branches in the value network to estimate $V_{\bm\phi}^{E}(\bm{s})$ and $V_{\bm\phi}^{E+I}(\bm{s})$. (b) AIRS selects the intrinsic reward function based on a queue (length=$W$) of estimated task return.}
		\label{fig:overview}
	\end{center}
	\vskip -0.2in
\end{figure*}

\subsection{Information Theory-Based Exploration}
Another representative idea is to design intrinsic rewards based on information theory \cite{seo2021state, mutti2021task, yuan2022renyi, yuan2022rewarding, yuan2022tackling}. \cite{houthooft2016vime} proposed to maximize the information gain about the agent’s belief of environment dynamics, which uses variational inference in Bayesian neural networks to handle continuous state and action spaces efficiently. \cite{kim2019emi} proposed to maximize the mutual information between related state-action representations and define the intrinsic reward using prediction error under a linear dynamics model. \cite{seo2021state} proposed to maximize the Shannon entropy of state visit distribution and designed a random-encoder-for-efficient-exploration (RE3) method. RE3 leverages a $k$-nearest neighbor estimator to estimate entropy efficiently and transforms the sample mean into particle-based intrinsic rewards. \cite{yuan2022renyi} further extended RE3 and proposed a R\'enyi state entropy maximization (RISE) method, which prevents the agent from visiting some states with a vanishing probability. In this paper, we selected RE3 and RISE as the candidates for the intrinsic reward set, providing powerful exploration incentives from the information theory perspective.

\section{Background}
\subsection{Preliminaries}
In this paper, we study the RL and control problems considering a  Markov Decision Process (MDP) \cite{bellman1957markovian} defined by a tuple $\mathcal{M}=\langle\mathcal{S},\mathcal{A},E,P,\rho(\bm{s}_{0}),\gamma\rangle$, where $\mathcal{S}$ is the state space, $\mathcal{A}$ is the action space, $E:\mathcal{S}\times\mathcal{A}\times\mathcal{S}\rightarrow\mathbb{R}$ is the extrinsic reward function that evaluates the actual task reward, $P(\bm{s}'|\bm{s},\bm{a}):\mathcal{S}\times\mathcal{A}\rightarrow\Delta(\mathcal{S})$ is the state-transition function that defines a probability distribution over $\mathcal{S}$, $\rho(\bm{s}_{0})$ is the initial state distribution, and $\gamma\in[0,1)$ is a discount factor. Denoting by $\pi_{\bm{\theta}}(\bm{a}|\bm{s})$ the policy of the agent, the objective of RL is to learn a policy that maximizes the expected discounted return $J_{E}(\bm{\theta})=\mathbb{E}_{\pi_{\bm{\theta}}}\left[\sum_{t=0}^{\infty}\gamma^{t}E_{t}\right]$ \cite{sutton2018reinforcement}.

In the following sections, we leverage intrinsic rewards to improve the exploration capability of the agent. Letting $I:\mathcal{S}\times\mathcal{A}\times\mathcal{S}\rightarrow\mathbb{R}$ denote the intrinsic reward function, and the optimization objective is redefined as
\begin{equation}
	J_{E+I}(\bm{\theta})=\mathbb{E}_{\pi_{\bm{\theta}}}\left[\sum_{t=0}^{\infty}\gamma^{t}(E_{t}+\beta_{t}\cdot I_{t})\right],
\end{equation}
where $\beta_{t}=\beta_{0}(1-\kappa)^{t}$ is a weighting coefficient that controls the degree of exploration, and $\kappa$ is a decay rate.

\subsection{Decoupled Advantage Actor-Critic}
Decoupled advantage actor-critic (DAAC) is a state-of-the-art on-policy algorithm that decouples the policy and value optimization for generalization in RL \cite{raileanu2021decoupling}. As shown in Figure~\ref{fig:overview}a, DAAC includes two separate networks, one for learning the policy and advantage and one for learning the state-value function. More specifically, the policy network of DAAC is trained to maximize the following objective:
\begin{equation}
	J_{\rm DAAC}(\bm{\theta})=J_{\pi}(\bm{\theta})+H_{\pi}(\bm{\theta})-L_{A}(\bm{\theta}),
\end{equation}
where $J_{\pi}$ is the policy gradient term of the proximal policy optimization (PPO) \cite{schulman2017proximal}, $H_{\pi}$ is an
entropy bonus to encourage exploration, and $L_{A}=[A_{\bm\theta}(\bm{s}_{t},\bm{a}_{t})-\hat{A}_{t}]^{2}$ is the advantage loss, and $\hat{A}_{t}$ is the corresponding generalized advantage estimate
at time step $t$ \cite{schulman2015high}. Finally, the value network of DAAC is trained to minimize the following loss:
\begin{equation}
	L_{V}(\bm{\phi})=\left[V_{\bm\phi}(\bm{s}_{t})-\hat{V}_{t}\right]^{2},
\end{equation}
where $\hat{V}_{t}$ is the total discounted return obtained during the corresponding episode after time step $t$.

\section{AIRS}
In this section, we propose the AIRS framework that improves the exploration and generalization ability of the RL agent by intelligently providing high-quality intrinsic rewards. Our critical insight is that different tasks and learning stages may benefit from distinct intrinsic reward functions. For instance, in the early stage of learning, aggressive exploration can enable agents to gain a lot of discriminative experiences in a short time. To that end, the simple idea is to design an intrinsic reward function that rewards significant state changes. However, in the later stage of learning, excessive exploration is unnecessary and may disturb the learned policy, and it suffices to keep appropriate exploration. The discussions above can be summarized as the following problems: 
\begin{itemize}
	\item Use or not use intrinsic reward function?
	\item Use which intrinsic reward function?
\end{itemize}

Denoting by $\mathcal{I}=\{I^{1},\dots,I^{n}\}$ the set of intrinsic reward functions, the problem of intrinsic reward selection can be formulated as a multi-armed bandit (Figure~\ref{fig:overview}b) \cite{lattimore2020bandit}. Each reward function is considered an arm, and the objective is to maximize the long-term return evaluated by the extrinsic reward function.

\subsection{Upper Confidence Bound}
Upper confidence bound (UCB) \cite{auer2002using} is an effective algorithm to solve the multi-armed bandit problem, which selects actions by the following policy:
\begin{equation}
	I_{t}=\underset{I\in\mathcal{I}}{\rm argmax}\bigg[Q_{t}(I)+c\sqrt{\frac{\log t}{N_{t}(I)}}\bigg],
\end{equation}
where $Q_{t}(I)$ is the estimated value of action $I$, $N_t(I)$ is the number of times that action $I$ has been selected prior to time $t$, and $c>0$ is a constant that controls the degree of exploration. To update $Q_{t}$, we employ two branches (Figure~\ref{fig:overview}a) in the value network to estimate the state-value function evaluated by the mixed reward function $E+I$ and extrinsic reward function $E$, respectively. For each intrinsic reward function $I$, a queue of length $W$ is leveraged to store the average estimated task return, and $Q_{t}(I)$ is computed as 
\begin{equation}\label{eq:queue estimation}
	Q_t(I)=\frac{1}{W}\sum_{i=1}^{W}\bar{V}_{\bm\phi}^{E}[i],
\end{equation}
where $\bar{V}_{\bm\phi}^{E}$ is the average estimated return of a rollout. Meanwhile, the loss function of the value network is redefined as
\begin{equation}
	\begin{aligned}
		L_{V}(\bm{\phi})&=\left[V_{\bm\phi}^{E+I}(\bm{s}_{t})-\hat{V}_{t}^{E+I}\right]^{2} \\
		&+\left[V_{\bm\phi}^{E}(\bm{s}_{t})-\hat{V}_{t}^{E}\right]^{2}.
	\end{aligned}
\end{equation}

Finally, the detailed workflow of AIRS based on DAAC is summarized in Algorithm~\ref{algo:airs on-policy}. We want to highlight two critical facts about AIRS. Here we use DAAC as the benchmark due to its improvement in generalization ability. But AIRS can be combined with many RL algorithms, and it suffices to use two branches in the value network to estimate task and total return, respectively. In addition, any appropriate bandit algorithms can replace the UCB algorithm. For example, the Thompson sampling \cite{russo2018tutorial} can be deployed when the intrinsic reward pool only has two reward functions. Therefore, AIRS is a very open architecture, and various attempts are possible.

\subsection{Intrinsic Reward Toolkit}
To facilitate the experiments and inspire subsequent research on intrinsic rewards, we developed a toolkit that provides high-quality implementations of diverse intrinsic reward modules based on PyTorch. This toolkit is designed to be highly modular and scalable. Each intrinsic reward module can be deployed in arbitrary algorithms in a plug-and-play manner, providing efficient and robust exploration increments. The following experiments are also performed based on this toolkit. More details about this toolkit can be found in Appendix~\ref{appendix:reward toolkit}.

\begin{algorithm*}[h]
	\caption{Automatic Intrinsic Reward Shaping (AIRS)}
	\label{algo:airs on-policy}
	\begin{algorithmic}[1]
		\State Initialize policy network $\pi_{\bm{\theta}}$ and value network $V_{\bm{\phi}}$ with parameters $\bm{\theta}$ and $\bm{\phi}$;
		\State Initialize the set of intrinsic reward functions $\mathcal{I}=\{I^{1},\dots,I^{n}\}$, exploration coefficient $c$, window length $W$ for estimating the Q-functions, and total number of updates $K$;
		\State $N(I)=1,\forall I\in\mathcal{I}$; \Comment{Initialize the number of times each $I$ was selected.}
		\State $Q(I)=0,\forall I\in\mathcal{I}$; \Comment{Initialize the Q-functions for all $I$.}
		\State $R(I)=\mathrm{FIFO}(W),\forall I\in\mathcal{I}$; \Comment{Initialize the lists of returns for all $I$.}
		\For{$k=1,\dots,K$}
		\State $I_{k}=\underset{I\in\mathcal{I}}{\rm argmax}\left[Q(I)+c\sqrt{\frac{\log k}{N(I)}}\right]$; \Comment{Use UCB to select an $I$.}
		\State Collect $\{(\bm{s}_{t},\bm{a}_{t},E_{t},\bm{s}_{t+1})\}_{t=1}^{T}$ using $\pi_{\bm{\theta}}$;
		\State \textcolor[RGB]{247,127,0}{Use $I_{k}$ to compute intrinsic rewards};
		\State Compute the value and advantage targets $\hat{V}^{E}_{t}, \hat{V}^{E+I}_{t}$ and $\hat{A}_{t}$ for all states $\bm{s}_{t}$;
		\State $\bm{\theta}\leftarrow \underset{\bm\theta}{\rm argmax}\: J_{\rm DAAC}$; \Comment{Update the policy network.}
		\State $\bm{\phi}\leftarrow \underset{\bm\phi}{\rm argmax}\: L_{V}$; \Comment{Update the value network.}
		\State \textcolor[RGB]{247,127,0}{Compute the mean return $\bar{V}_{\bm\phi}$ obtained by the new policy};
		\State Add $\bar{V}_{\bm\phi}$ to the $R(I_{k})$ list using the first-in-first-out rule;
		\State $Q(I_{k})\leftarrow \frac{1}{|R(I_{k})|}\sum_{\bar{V}_{\bm\phi}\in R(I_k)} \bar{V}_{\bm\phi}$;
		\State $N(I_{k})\leftarrow N(I_{k})+1$.
		\EndFor
	\end{algorithmic}
\end{algorithm*}

\begin{figure*}[h!]
	\vskip 0.2in
	\begin{center}
		\centerline{\includegraphics[width=\linewidth]{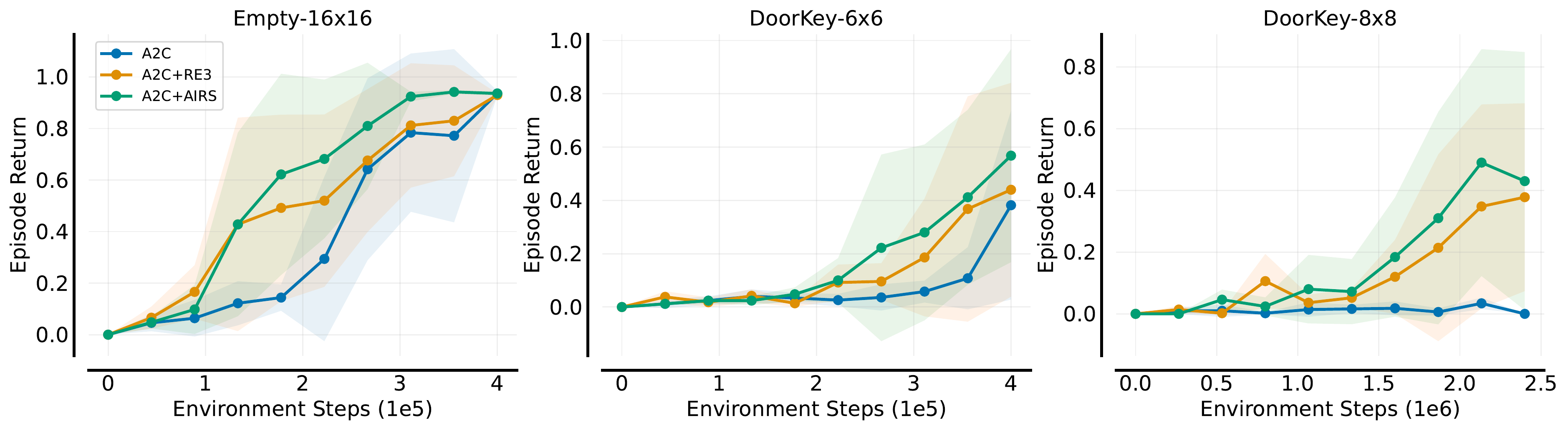}}
		\caption{Sample efficiency comparison between AIRS and benchmarks. The solid line and shaded regions represent the mean and standard deviation, which are computed over five random seeds.}
		\label{fig:minigrid_sample_efficiency}
	\end{center}
	\vskip -0.2in
\end{figure*}

\begin{figure*}
	\vskip 0.2in
	\begin{center}
		\centerline{\includegraphics[width=\linewidth]{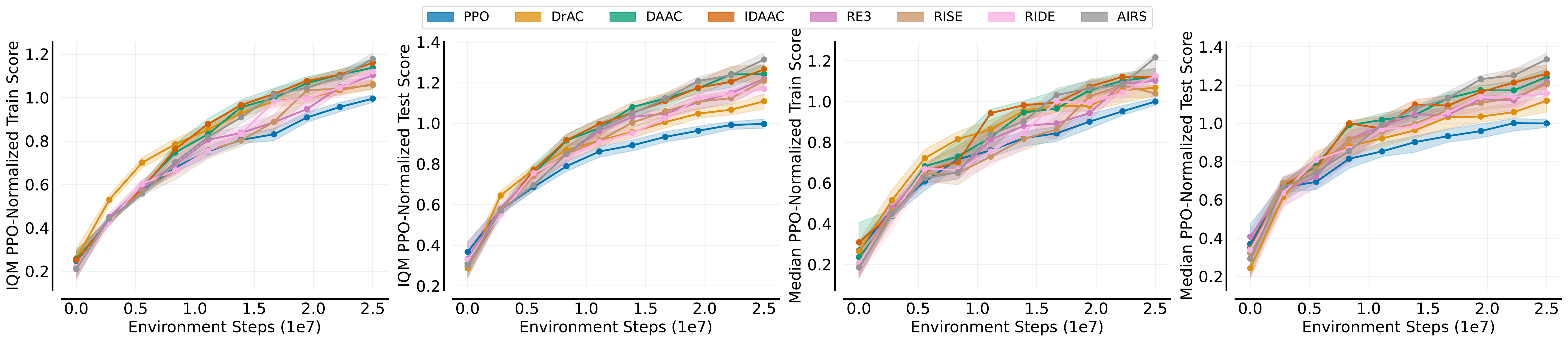}}
		\caption{Comparing Median vs IQM on train and test levels. Sample-efficiency of agents as a function of number of frames measured via \textbf{IQM} (two on the left) and \textbf{Median} (two on the right) PPO-Normalized scores. Shaded regions show pointwise 95\% percentile stratified bootstrap confidence intervals.}
		\label{fig:procgen_sample_efficiency}
	\end{center}
	\vskip -0.2in
\end{figure*}

\begin{table*}[h]
	\centering
	\caption{Four intrinsic reward functions are employed in the Procgen experiments, where $\bm{e}=g(\bm{s})$ is the representation of $\bm{s}$, $N_{\rm ep}$ is the episodic state visitation frequency. For RE3 and RISE, $g$ is a random and fixed encoder, while $g$ is a learned encoder for RIDE.  $\tilde{\bm{e}}$ is the $k$-nearest neighbor of $\bm{s}$ in the encoding space. More details about the implementations can be found in Appendix~\ref{appendix:procgen experiments}.}
	\label{tb:irs list}
	\vskip 0.15in
	\begin{tabular}{lll}
		\toprule
		\textbf{Intrinsic reward module} & \textbf{Formulation} & \textbf{Remark}\\
		\midrule
		RE3 \cite{seo2021state}    & $I_{t}=\frac{1}{k}\sum_{i=1}^{k}\log(\Vert\bm{e}_{t}-\tilde{\bm{e}}_{t}^{i}\Vert_{2}+1)$  & Shannon entropy maximization\\ \midrule
		RISE \cite{yuan2022renyi}  &  
		$I_{t}=\frac{1}{k}\sum_{i=1}^{k}(\Vert\bm{e}_{t}-\tilde{\bm{e}}_{t}^{i}\Vert_{2})^{1-\alpha}$ & R\'enyi entropy maximization \\ \midrule
		RIDE \cite{raileanu2020ride} & 
		$I_{t}=\Vert\bm{e}_{t+1}-\bm{e}_{t}\Vert_{2}/\sqrt{N_{\rm ep}(\bm{s}_{t+1})}$
		& Significant state changes \\ \midrule
		ID & $I_{t}=0$ & No shaping \\
		\bottomrule
	\end{tabular}
	%	\vskip -0.1in
\end{table*}

\section{Experiments}
In this section, we designed experiments to answer the following questions:
\begin{itemize}
	\item Can AIRS alleviate the aforementioned biased objective problem? (See Figure~\ref{fig:minigrid_sample_efficiency})
	\item Can AIRS improve policy performance in both continuous and discrete control tasks? (See Table~\ref{tb:procgen results train}, Table~\ref{tb:procgen results test}, Figure~\ref{fig:minigrid_sample_efficiency}, and Figure~\ref{fig:dmc_sample_efficiency})
	\item How does AIRS compare to single intrinsic reward-driven approaches? (See Figure~\ref{fig:procgen_sample_efficiency}, Figure~\ref{fig:procgen_test_aggregates}, Figure~\ref{fig:procgen_test_poi}, Table~\ref{tb:procgen results train}, and Table~\ref{tb:procgen results test})
	\item Can the redesigned value network make effective value estimation? (See Figure~\ref{fig:procgen value loss full})
	\item What is the detailed decision process of AIRS during training? (See Figure~\ref{fig:procgen_selection_part})
\end{itemize}

In particular, we report all the experiment results using a reasonable and reliable toolbox entitled \href{https://github.com/google-research/rliable}{rliable} to guarantee reliability and reduce the uncertainty \cite{agarwal2021deep}. Four metrics are introduced in the following sections, \textbf{Median}: Aggregate median performance, higher is better. \textbf{IQM}: Aggregate interquartile mean performance, higher is better. \textbf{Mean}: Aggregate mean performance, higher is better. \textbf{Optimality Gap}: The amount by which the algorithm fails to meet a minimum score, lower is better.

\subsection{MiniGrid Games}
\subsubsection{Setup}
We first tested AIRS on the MiniGrid benchmark to highlight the effectiveness of automatic intrinsic reward shaping \cite{chevalier2018Minimalistic}. MiniGrid contains a collection of 2D grid-world environments with goal-oriented and sparse-reward tasks that are extremely hard-exploration. We selected advantage actor-critic (A2C) \cite{mnih2016asynchronous} as the baseline and considered the combination of A2C and RE3. For AIRS, we build it on top of A2C+RE3, in which the intrinsic reward function pool only has ID and RE3. See Table~\ref{tb:irs list} for the details of the two functions. In particular, we added the intrinsic rewards into estimated advantages directly rather than using an additional branch in the value network. For each benchmark, we only report the best results after a hyper-parameter search, and more details on MiniGrid experiments are provided in Appendix~\ref{appendix:minigrid experiments}.

\subsubsection{Results}
Figure~\ref{fig:minigrid_sample_efficiency} illustrates the sample efficiency comparison between AIRS and benchmarking schemes. A2C+RE3 successfully outperformed the vanilla A2C agent in all three games, demonstrating that RE3 can effectively promote the sample efficiency using the same environment steps. In contrast, A2C+AIRS achieves the highest performance and convergence rate in all three games. This indicates that AIRS successfully alleviated the biased objective problem and controlled the exploration degree intelligently in the training process.

\begin{figure*}
	\vskip 0.2in
	\begin{center}
		\centerline{\includegraphics[width=\linewidth]{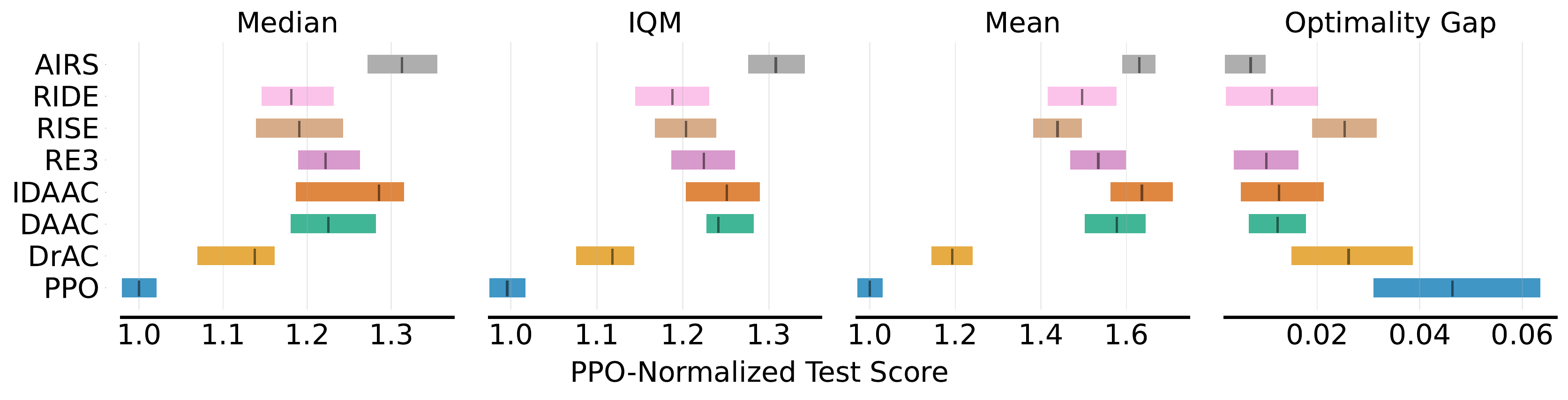}}
		\caption{PPO-Normalized test performance for the Procgen benchmark, which are aggregated over all sixteen tasks and ten randoms seeds.}
		\label{fig:procgen_test_aggregates}
	\end{center}
	\vskip -0.2in
\end{figure*}

\begin{figure}
	\vskip 0.2in
	\begin{center}
		\centerline{\includegraphics[width=\linewidth]{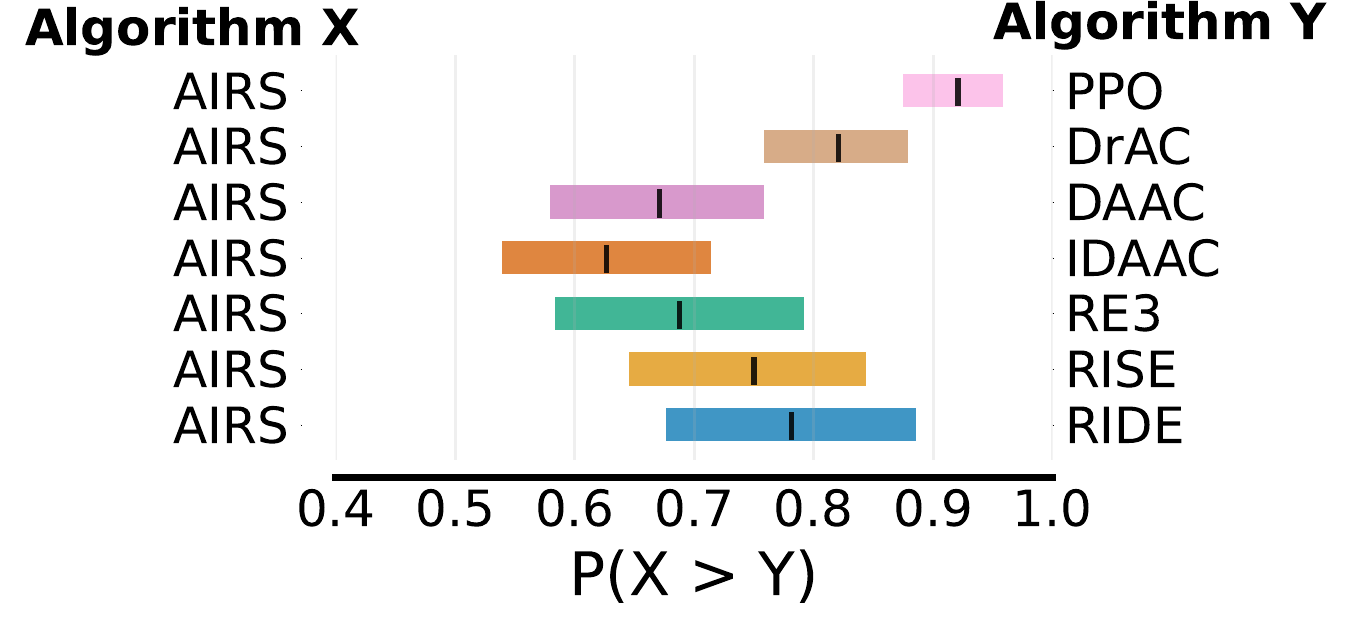}}
		\caption{Each row shows the probability of
			improvement, with 95\% bootstrap confidence intervals, that the algorithm $X$ on the left outperforms algorithm $Y$ on the right,
			given that X was claimed to be better than $Y$. For all algorithms, results are based on ten random seeds per task.}
		\label{fig:procgen_test_poi}
	\end{center}
	\vskip -0.2in
\end{figure}

\subsection{Procgen Games}
\subsubsection{Setup}
Next, we tested AIRS on the full Procgen benchmark containing sixteen games with procedurally-generated environments \cite{cobbe2020leveraging}. Procgen is developed similarly to the ALE benchmark, in which the agents have to learn motor control directly from images \cite{bellemare2013arcade}. But Procgen has higher requirements for the generalization ability, and the agent has to make sufficient exploration and learn transferable skills rather than memorizing specific trajectories. These attributes make it a good benchmark for our research. All Procgen games utilize a discrete action space with 15 possible actions and produce $64\times64\times3$ RGB observations. The agents were trained on the easy level and tested on the entire distribution of levels.

To construct the intrinsic reward set, four intrinsic reward functions were selected to serve as the candidates, namely RE3, RISE, RIDE, and ID, respectively \cite{seo2021state, yuan2022renyi, raileanu2020ride}. The detailed information of these functions is illustrated in Table~\ref{tb:irs list}. The reasons for selection are as follows: (i) they can provide sustainable exploration incentives, {\em i.e.}, the intrinsic rewards will not vanish with visits; (ii) they are computation-efficeint and easy to implement, and (iii) they can leverage fixed and random encoder to encode the state space, which provides relatively stable intrinsic reward space and guarantees the algorithm convergence.

Furthermore, we selected data-regularized actor-critic (DrAC), IDAAC, DAAC, and PPO to serve as the benchmarks \cite{raileanu2020automatic, raileanu2021decoupling, schulman2017proximal}. DrAC combines data augmentation and actor-critic algorithms theoretically, demonstrating remarkable performance on Procgen benchmark. IDAAC is a variant of DAAC that adds an additional regularizer to the DAAC policy encoder to guarantee that it only contains episode-specific information. For each benchmark method, we only report its best performance after a hyper-parameter search, and more details on Prcogen games are provided in Appendix~\ref{appendix:procgen experiments}.

\subsubsection{Results}
The game complexity of test levels is much higher than the train levels, which provides a direct measure to evaluate the generalization ability of the agents. Figure~\ref{fig:procgen_test_aggregates} illustrates the test performance comparison evaluated by four metrics. AIRS achieved the highest performance in all four metrics, especially for the \textbf{Median} and \textbf{IQM}. In contrast, IDAAC achieved the second highest performance in all four metrics. The performance of IDAAC and DAAC is very close in most games. This indicated that the additional regularization term did not significantly improve the performance of DAAC, which was the same as the results reported in \cite{raileanu2021decoupling}. Moreover, the combination of DAAC and single intrinsic rewards achieved lower performance than the vanilla DAAC agent. Furthermore, we computed the probability of improvement of AIRS over the benchmarks. Figure~\ref{fig:procgen_test_poi} indicates that the probability of AIRS being better than DAAC is 67\%, and the probability of AIRS being better than IDAAC is 61\%. We also provided a complete list of final performance comparisons in Table~\ref{tb:procgen results train} and Table~\ref{tb:procgen results test}.

\begin{figure*}[h!]
	\vskip 0.2in
	\begin{center}
		\centerline{\includegraphics[width=\linewidth]{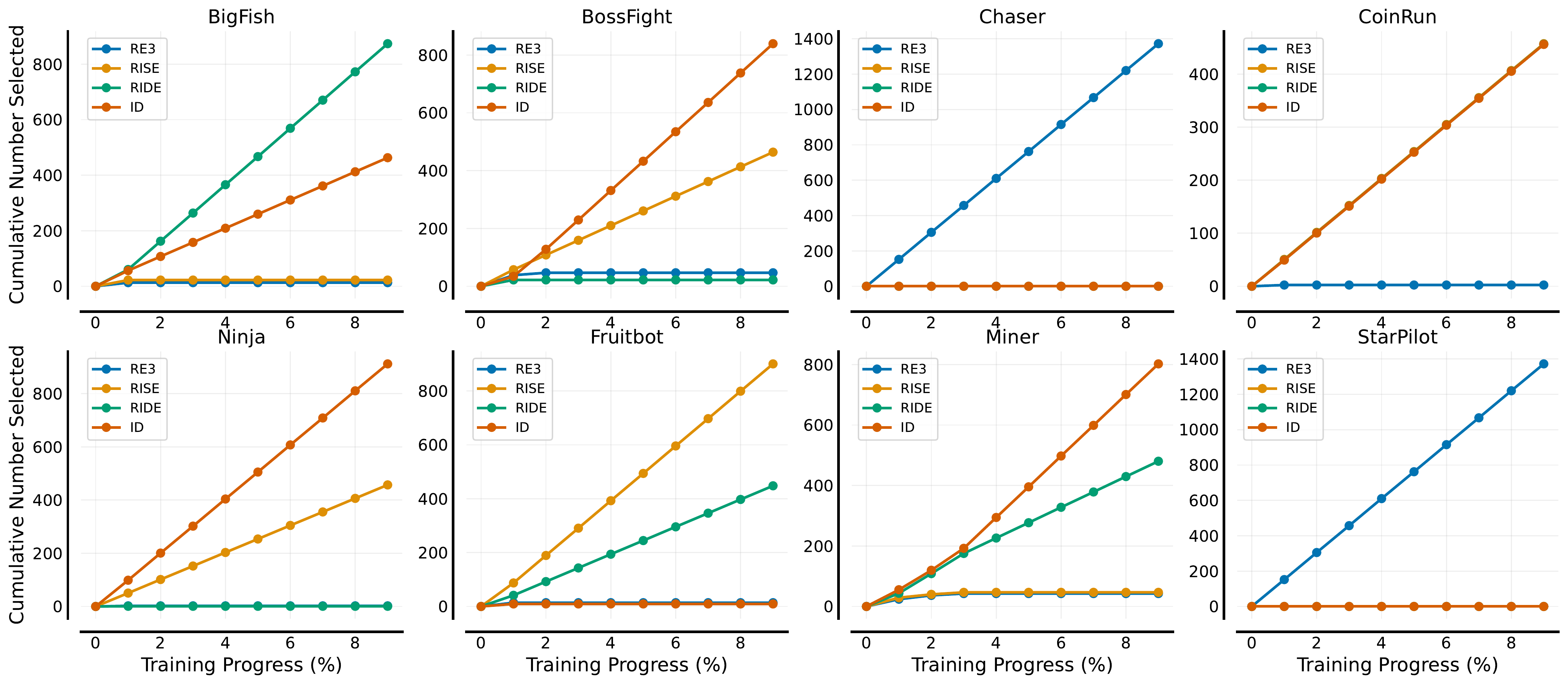}}
		\caption{The cumulative number of times AIRS selects each intrinsic reward function over the training progress, computed with ten random seeds.}
		\label{fig:procgen_selection_part}
	\end{center}
	\vskip -0.2in
\end{figure*}

\begin{figure*}[h!]
	\vskip 0.2in
	\begin{center}
		\centerline{\includegraphics[width=\linewidth]{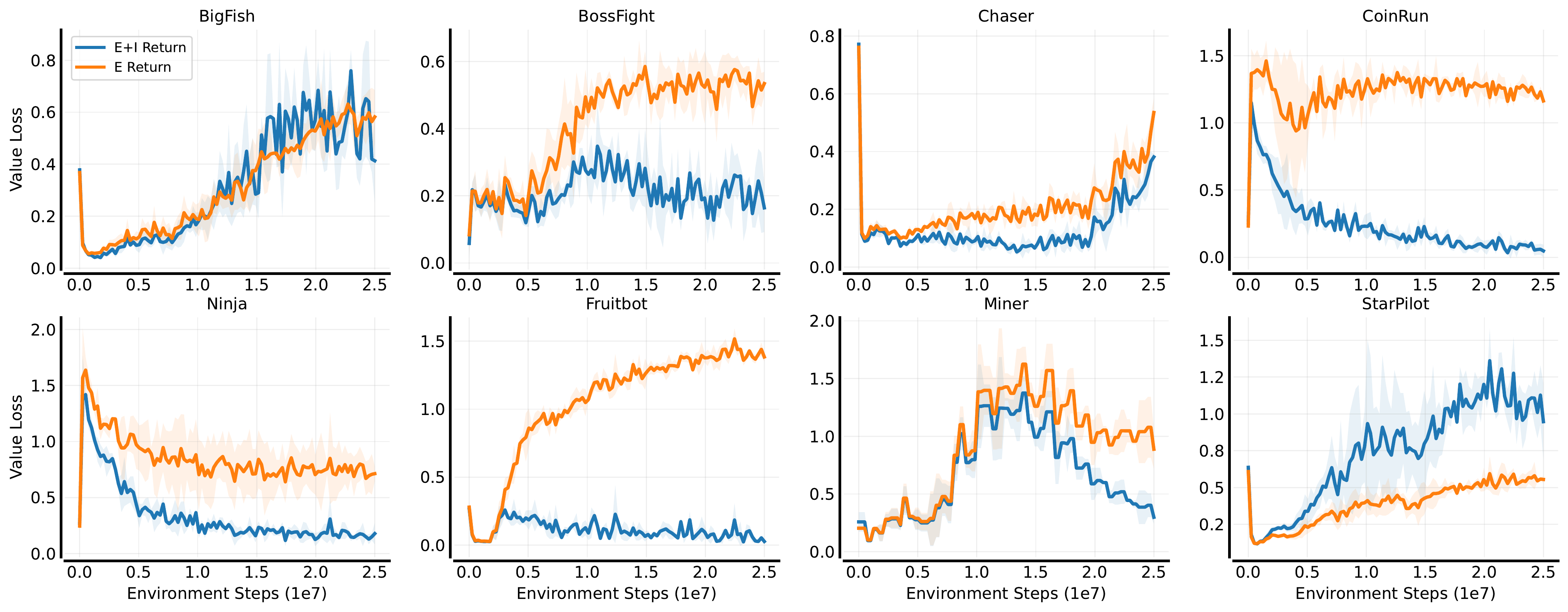}}
		\caption{Value loss of eight selected games during training. "E Return" indicates the branch for extrinsic value estimation in the value network. Full loss curves can be found in in Appendix A. The solid line and shaded regions represent the mean and standard deviation over 10 random seeds, respectively.}
		\label{fig:procgen_value_loss_part}
	\end{center}
	\vskip -0.2in
\end{figure*}

Figure~\ref{fig:procgen_sample_efficiency} illustrates the sample efficiency comparison between AIRS and benchmarking schemes on the full Procgen benchmark. AIRS had a lower convergence rate in the early stage, which was mainly limited by the estimation accuracy of the value network. Figure~\ref{fig:procgen_value_loss_part} illustrates the value loss of eight games during training, and full loss curves can be found in Appendix~\ref{appendix:procgen experiments}. It is evident that the loss of the two branches tended to stabilize after 1M environment steps. But AIRS still achieved remarkable performance gains. This indicates that AIRS can effectively improve the exploration of the agent and keep tracking the fundamental objective, {\em i.e.}, the task reward.

Furthermore, we documented the specific decision-making process of AIRS. Figure~\ref{fig:procgen_selection_part} illustrates the cumulative number of times AIRS selects each intrinsic reward function over the training progress of eight games, and full curves can be found in Appendix~\ref{appendix:procgen experiments}. In \textit{BigFish} game, the RIDE was the most selected function while AIRS chose no shaping in $33.7\%$ of the training time. In \textit{Miner} game, the RISE was the most selected function while AIRS chose no shaping in $59.3\%$ of the training time. Similarly, the RISE was also the most selected function in \textit{Ninja} game, and AIRS chose no shaping in $66.5\%$ of the training time. Meanwhile, AIRS divided the choice opportunity into three functions in \textit{CoinRun} game and \textit{BossFight} game. It is evident that no shaping played an essential role in the whole training process, and appropriate exploration can improve sample efficiency and gain higher performance. In contrast, AIRS selected RISE and RIDE at most in \textit{Fruitbot} game. In \textit{Chaser} and \textit{StarPilot} game, AIRS only selected the RE3 function during the training, and the final performance of AIRS and DAAC+RE3 is almost the same.

\subsubsection{Ablations}
Like the MiniGrid experiments, we performed an ablation study by setting the intrinsic reward pool to have only two modules: RE3 and ID. As a result, the test \textbf{IQM} and \textbf{OG} are 1.27 and 0.015, which was better than DAAC+RE3. The ablation results further proved that AIRS can effectively alleviate the bias objective problem, providing appropriate exploration incentives when necessary. It also demonstrated that AIRS could assemble advantages by automatically selecting from multiple intrinsic rewards.

\begin{figure*}
	\vskip 0.2in
	\begin{center}
		\centerline{\includegraphics[width=1.\linewidth]{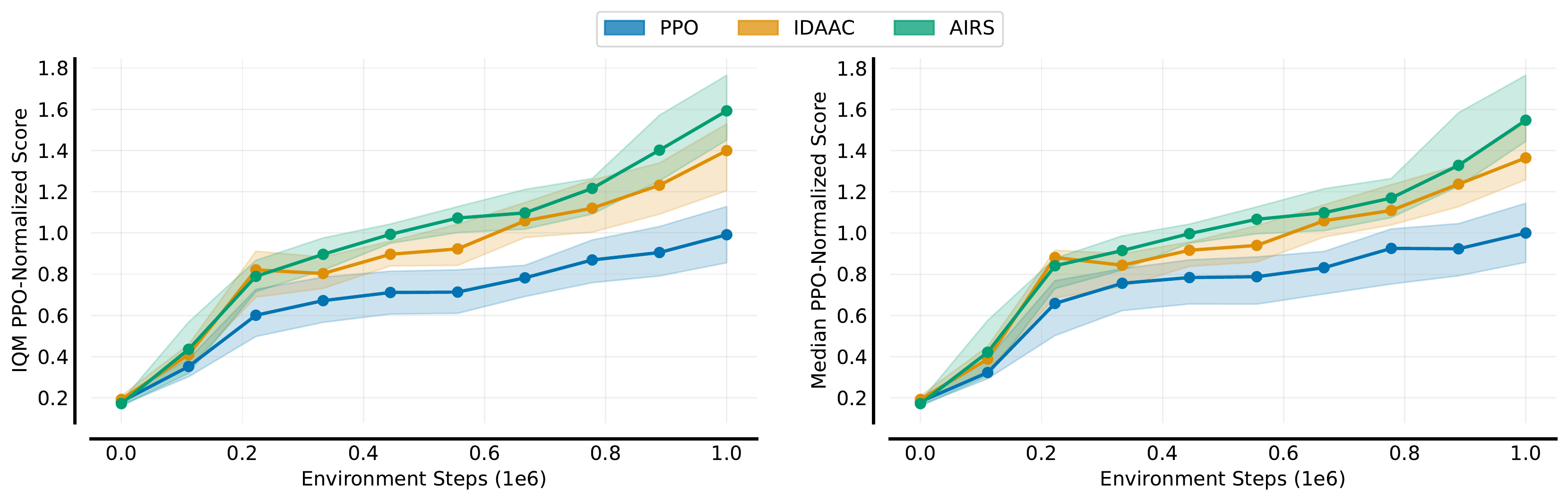}}
		\caption{Sample-efficiency of agents as a function of number of frames measured via \textbf{IQM} and \textbf{Median} PPO-Normalized scores. Shaded regions show pointwise 95\% percentile stratified bootstrap confidence intervals.}
		\label{fig:dmc_sample_efficiency}
	\end{center}
	\vskip -0.2in
\end{figure*}

\subsection{DeepMind Control Suite}
\subsubsection{Setup}
Finally, we tested AIRS on the DeepMind Control Suite with distractors, and four tasks were introduced, namely \textit{Cartpole Balance}, \textit{Cheetah Run}, \textit{Finger Spin}, and \textit{Walker Walk}, respectively \cite{tassa2018deepmind}. For each task, the backgrounds of image observations are replaced using natural videos from the Kinetics dataset \cite{cobbe2019quantifying}. Note that the background is sampled from a list of videos at the beginning of each episode, which creates spurious correlations between backgrounds and rewards. For each benchmark, we only report the best results after a hyper-parameter search, and more details on DeepMind Control Suite experiments are provided in Appendix~\ref{appendix:dmc experiments}.

\subsubsection{Results}
Figure~\ref{fig:dmc_sample_efficiency} illustrates the sample efficiency comparison between AIRS and benchmarking schemes after training 500K frames. AIRS achieved the highest performance in all the tasks, while IDAAC outperformed the vanilla PPO agent in all the tasks. Therefore, AIRS can improve the performance of RL agents on both discrete and continuous control tasks.

\section{Discussion}
In this paper, we investigate the problem of enhancing exploration in RL and propose an intrinsic reward-driven method entitled AIRS. AIRS can intelligently select the best intrinsic reward function from a pre-defined set in real-time, providing reliable exploration incentives and guaranteeing policy invariance. In particular, we develop a toolkit to provide high-quality implementations of diverse intrinsic reward approaches, which is expected to inspire more subsequent research. We test AIRS on multiple tasks from MiniGrid, Procgen, and DeepMind Control Suite. Extensive simulation results demonstrate that our method can achieve superior performance. 

However, there are some remaining limitations of AIRS. Firstly, AIRS performs selection based on a pre-defined reward set. The quality of selected reward modules will inevitably affect the final policy performance. It's critical to find an effective method to build the reward set. Meanwhile, different intrinsic reward spaces may interfere with each other, resulting in unexpected learning collapse. Secondly, AIRS makes decisions based on the estimated task return, which requires an independent module to make estimations. The estimation accuracy also determines the reliability of actions of AIRS, and additional modules will increase the complexity of the deployment of AIRS. For algorithms like A2C and PPO, a simple method is to add the intrinsic rewards into estimated advantages directly rather than using an additional branch in the value network, which is used in our MiniGrid experiments. Finally, AIRS formulates the reward selection as a bandit problem and uses UCB to solve it, in which enough attempts are required to estimate the bound. This will significantly increase the computational overhead and decrease the convergence rate. We will address these issues in future work.

\section*{Acknowledgements}
This work was supported, in part, by NSFC under Grant No. 62102333, HKSAR RGC under Grant No. PolyU 25211321, and ZJNSFC under Grant LQ23F010008. We also sincerely thank all the reviewers and editors for their hard work and insightful comments, which dramatically enhances the quality of our manuscript.

\clearpage
%\bibliography{reference}
%\bibliographystyle{icml2023}

%%%%%%%%%%%%%%%%%%%%%%%%%%%%%%%%%%%%%%%%%%%%%%%%%%%%%%%%%%%%%%%%%%%%%%%%%%%%%%%
%%%%%%%%%%%%%%%%%%%%%%%%%%%%%%%%%%%%%%%%%%%%%%%%%%%%%%%%%%%%%%%%%%%%%%%%%%%%%%%
% APPENDIX
%%%%%%%%%%%%%%%%%%%%%%%%%%%%%%%%%%%%%%%%%%%%%%%%%%%%%%%%%%%%%%%%%%%%%%%%%%%%%%%
%%%%%%%%%%%%%%%%%%%%%%%%%%%%%%%%%%%%%%%%%%%%%%%%%%%%%%%%%%%%%%%%%%%%%%%%%%%%%%%
\newpage
\appendix
\onecolumn
\section{Details on MiniGrid Experiments}\label{appendix:minigrid experiments}
\subsection{Environment Setting}
\begin{figure*}[h]
	\vskip 0.2in
	\begin{center}
		\centerline{\includegraphics[width=0.95\linewidth]{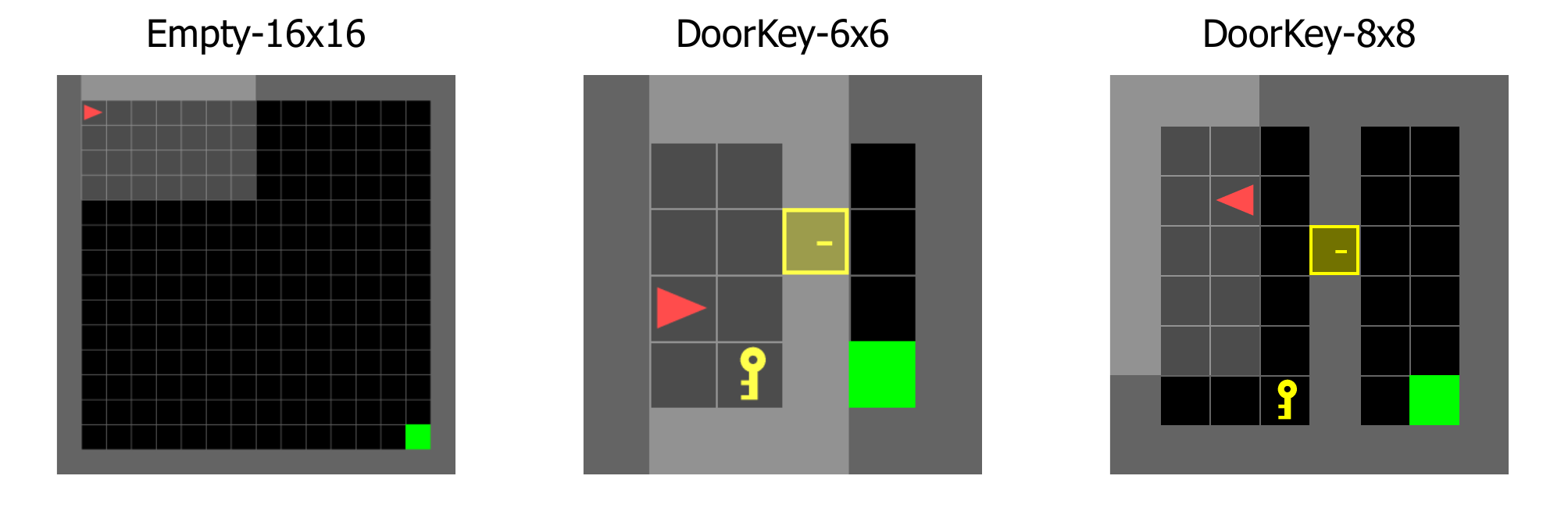}}
		\caption{Screenshots of three MiniGrid games.}
		\label{fig:minigrid screenshots}
	\end{center}
	\vskip -0.2in
\end{figure*}

In this section, we evaluated the performance of AIRS on three tasks from MiniGrid games \cite{chevalier2018Minimalistic}, namely \textit{Empty-16x16}, \textit{DoorKey-6x6}, and \textit{DoorKey-8x8}, respectively. The environment code can be found in the publicly available released repository (\url{https://github.com/Farama-Foundation/MiniGrid}). Figure~\ref{fig:minigrid screenshots} illustrates the screenshots of the three games. The girdworld was set to be fully observable, and a compact grid encoding was used to generate observations (7$\times$7$\times$3).

\subsection{Experiment Setting}
\textbf{A2C}. \cite{mnih2016asynchronous} For A2C, we followed the implementation in the publicly released repository (\url{https://github.com/lcswillems/rl-starter-files}), and used their default hyperparameters listed in Table~\ref{tb:minigrid hyperparameters}.

\textbf{A2C+RE3}. \citep{seo2021state} For RE3, we followed the implementation in the publicly released repository (\url{https:
	//github.com/younggyoseo/RE3}). Here, the intrinsic reward is computed as $I_{t}=\log(\Vert\bm{e}_{t}-\tilde{\bm{e}}_{t}\Vert_{2}+1)$, where $\bm{e}_{t}=g(\bm{s}_{t})$ and $g$ is a random and fixed encoder. The total reward of time step $t$ is computed as $E_{t}+\beta_{t}\cdot I_{t}$, where $\beta_{t}=\beta_{0}(1-\kappa)^{t}$. Moreover, the average distance of $\bm{e}_t$ and its $k$-nearest neighbors was used to replace the single $k$ nearest neighbor to provide a less noisy state entropy estimate. As for hyperparameters related to exploration, we used $k=3, \kappa=0$ for all three games, $\beta_{0}=0.1$ for Empty-16x16, $\beta_{0}=0.005$ for DoorKey-6x6, and $\beta_{0}=0.01$ for DoorKey-8x8, respectively.

\textbf{A2C+AIRS}. For AIRS, we built it on top of A2C+RE3, which uses same policy and value network architectures with A2C and A2C+RE3. We performed a hyperparameter search over the initial exploration degree $\beta_{0}\in\{0.005, 0.01, 0.1\}$, the decay rate $\kappa\in\{0.0, 0.00001,0.000025,0.00005\}$, the exploration coefficient $c\in\{0.0, 0.1, 0.5, 1.0, 5.0\}$ and the size of sliding window used to compute the Q-values $W\in\{10, 50, 100\}$. We found that the best values are $\beta_{0}=0.1$ for Empty-16x16, $\beta_{0}=0.005$ for DoorKey-6x6, $\beta_{0}=0.01$ for DoorKey-8x8, $\kappa=0$, $c=0.1$, and $W=10$, which were used to obtain the results reported here.

\begin{table}[h]
	\centering
	\caption{General hyperparameters used to obtain the MiniGrid results.}
	\label{tb:minigrid hyperparameters}
	\vskip 0.15in
	\begin{tabular}{ll}
		\toprule
		\textbf{Hyperparameter}             & \textbf{Value}         \\ \midrule
		Observation downsampling   & (7, 7, 3)      \\
		Stacked frames             & No            \\
		Environment steps          & 400000 Empty-16x16, 400000 DoorKey-6x6, 2400000 DoorKey-8x8     \\
		Episode steps 			   & 5           \\
		Number of workers          & 1             \\
		Environments per worker    & 16            \\
		Optimizer                  & RMSprop          \\
		Learning rate              & 0.001        \\
		GAE coefficient            & 0.95          \\
		Action entropy coefficient & 0.01          \\
		Value loss coefficient     & 0.5           \\
		Value clip range           & 0.2           \\
		RMSprop $\epsilon$         & 0.01          \\
		Max gradient norm          & 0.5           \\
		Epochs per rollout         & 3             \\
		Mini-batches per epoch     & 8             \\
		LSTM                       & No            \\
		Gamma $\gamma$             & 0.99          \\ \bottomrule
	\end{tabular}
	\vskip -0.1in
\end{table}

\clearpage
\section{Details on Procgen Experiments}\label{appendix:procgen experiments}
\subsection{Environment Setting}
\begin{figure*}[h]
	\vskip 0.2in
	\begin{center}
		\centerline{\includegraphics[width=0.97\linewidth]{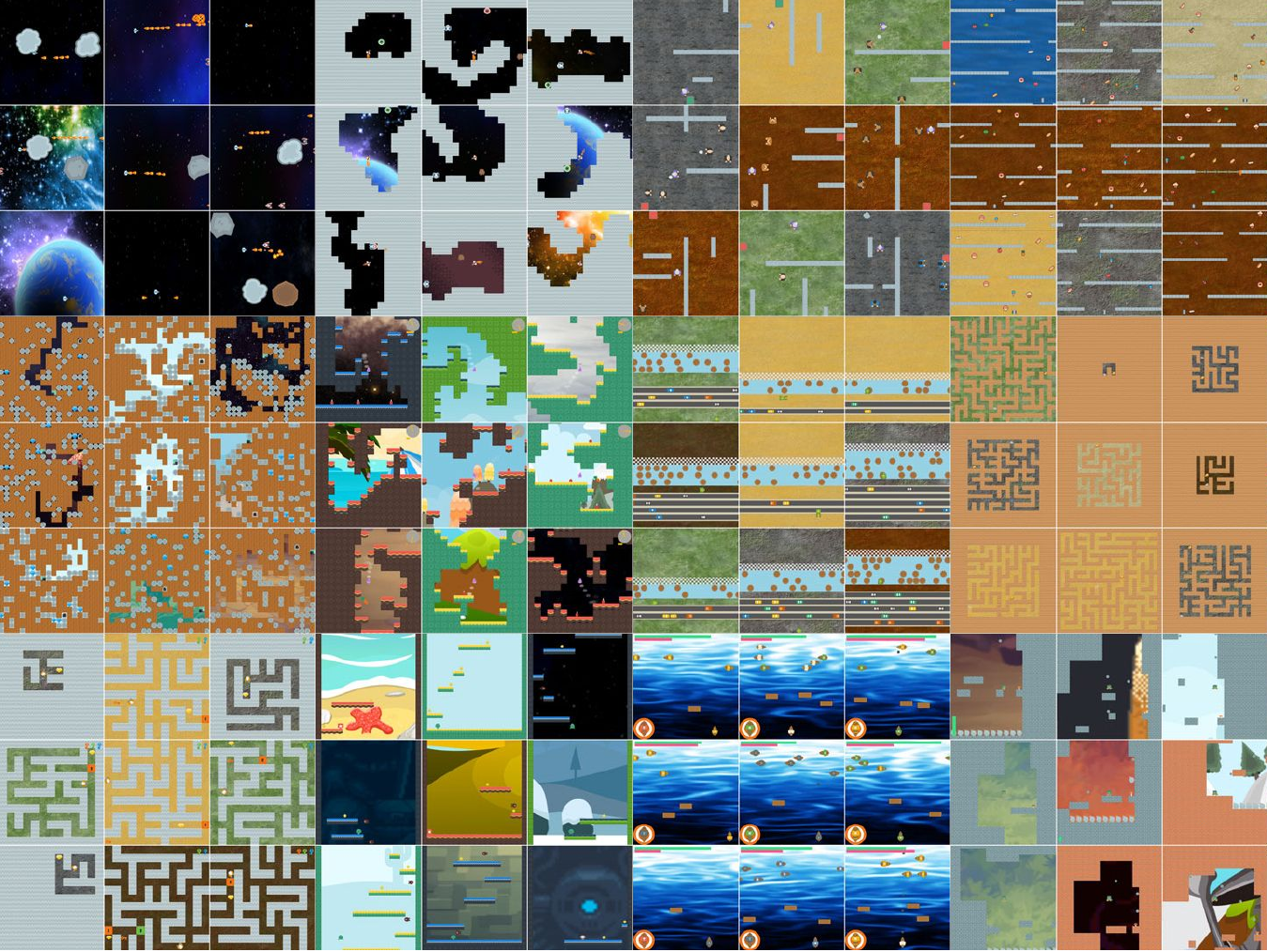}}
		\caption{Screenshots of multiple procedurally-generated levels from 12 Procgen environments: StarPilot, CaveFlyer, Dodgeball, FruitBot, Miner, Jumper, Leaper, Maze, Heist, Climber, Plunder, Ninja (from left to right, top to bottom).}
		\label{fig:procgen screenshots}
	\end{center}
	\vskip -0.2in
\end{figure*}

In this section, we evaluated the performance of AIRS on the full Procgen benchmark \cite{cobbe2020leveraging}. All Procgen games utilize a discrete action space with 15 possible actions and produce $64\times64\times3$ RGB observations. Figure~\ref{fig:procgen screenshots} illustrates the screenshots of multiple procedurally-generated levels from 12 Procgen environments. The environment code can be found in the publicly available released repository (\url{https://github.com/openai/procgen}). The agents were trained on a fixed set of 200 levels ({\em i.e.}, generated using seeds from 1 to 200) and tested on the full distribution of levels ({\em i.e.}, generated using randomly-sampled computer integers).

\subsection{Experiment Setting}

\begin{figure*}[h]
	\vskip 0.2in
	\begin{center}
		\centerline{\includegraphics[width=0.85\linewidth]{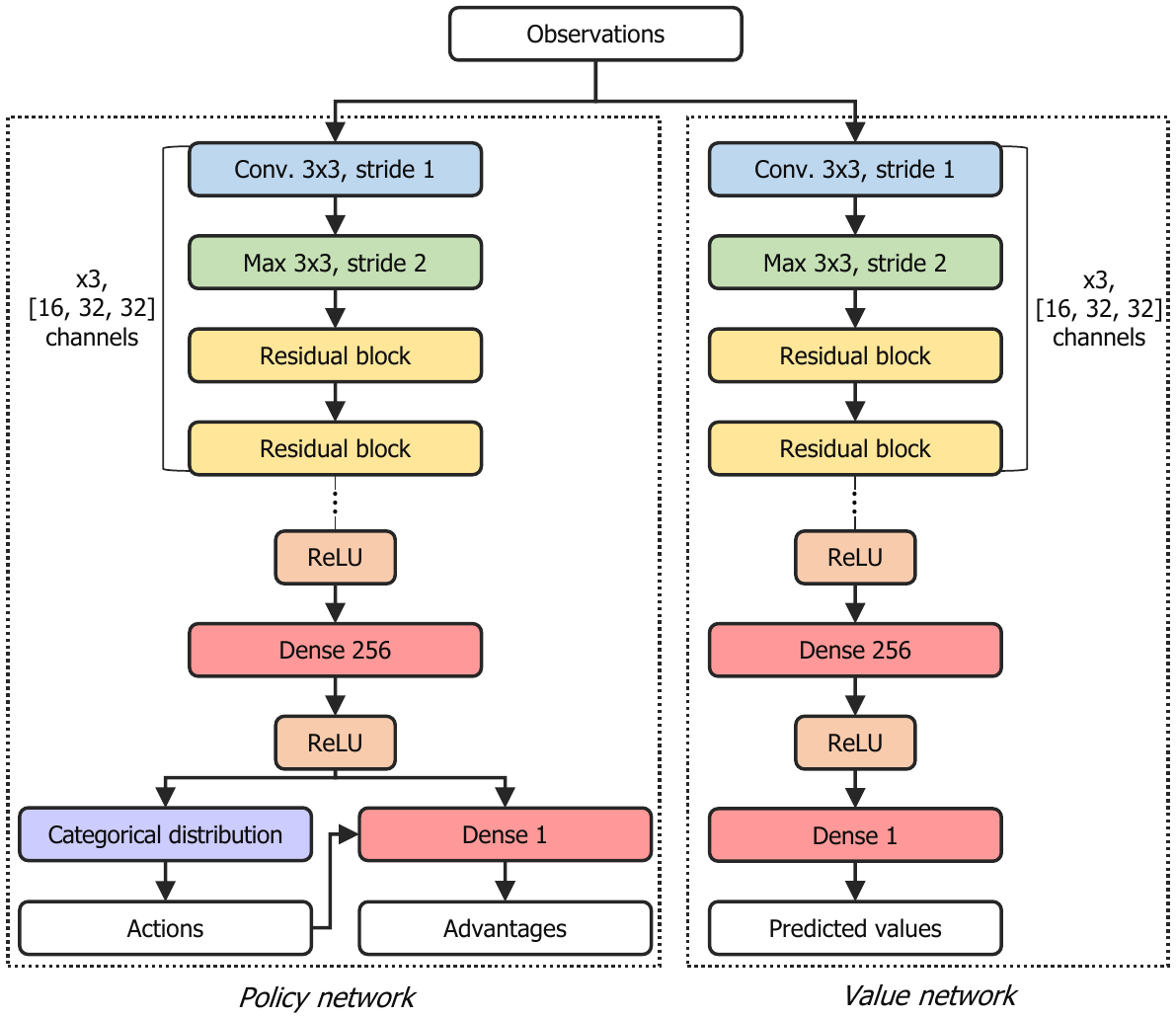}}
		\caption{ResNet-based architectures of the policy network and the value network. Here the residual block consists of two convolutional layers with ReLU function \cite{glorot2011deep, he2016deep}.}
		\label{fig:network}
	\end{center}
	\vskip -0.2in
\end{figure*}

\textbf{AIRS}. In this experiment, we built AIRS on top of DAAC \cite{raileanu2021decoupling} and employed two branches in the value network to predict the state-value function evaluated by the mixed reward function and extrinsic reward function, respectively. Figure~\ref{fig:network} illustrates the ResNet-based architectures of the policy network and the value network \cite{espeholt2018impala}. For each game, the agents were trained for 25M frames on the easy mode and tested using the entire
distribution of levels. 

We performed a hyperparameter search over the initial exploration degree $\beta_{0}\in\{0.01, 0.05, 0.1\}$, the decay rate $\kappa\in\{0.00001,0.000025,0.00005\}$, the exploration coefficient $c\in\{0.0, 0.1, 0.5, 1.0, 5.0\}$ and the size of sliding window used to compute the Q-values $W\in\{10, 50, 100\}$. We found that the best values are $\beta_{0}=0.05$, $\kappa=0.000025$, $c=0.1$, and $W=10$, which were used to obtain the results reported here.

\textbf{PPO}. \cite{schulman2017proximal} For PPO, we followed the implementation in the publicly released repository (\url{https://github.com/ikostrikov/pytorch-a2c-ppo-acktr-gail}). To make a fair comparison, the agent was parameterized by the same ResNet architecture in Figure~\ref{fig:network} to obtain the best results. More detailed hyperparameters are illustrated in Table~\ref{tb:procgen hyperparameters}.

\textbf{DrAC}. \cite{raileanu2020automatic} For DrAC, we followed the implementation in the publicly released repository (\url{https://github.com/rraileanu/auto-drac}). DrAC was implemented based on the PPO repository above and added an augmentation loss term into the original PPO loss function. For each game, we used the best augmentation type reported in \cite{raileanu2020automatic}, which is illustrated in Table~\ref{tb:drac aug type}. The coefficient of the augmentation loss was $0.1$, and the remaining hyperparameters were the same as Table~\ref{tb:procgen hyperparameters}.

\begin{table}[]
	\centering
	\caption{Augmentation type used for each game in DrAC.}
	\label{tb:drac aug type}
	\small
	\begin{tabular}{lllllllll}
		\toprule 
		\textbf{Game}         & BigFish     & StarPilot & FruitBot     & BossFight & Ninja        & Plunder & CaveFlyer & CoinRun      \\ \midrule
		\textbf{Augmentation} & crop        & crop      & crop         & flip      & color-jitter & crop    & rotate    & random-conv  \\ \midrule
		\textbf{Game}         & Jumper      & Chaser    & Climber      & Dodgeball & Heist        & Leaper  & Maze      & Miner        \\ \midrule
		\textbf{Augmentation} & random-conv & crop      & color-jitter & crop      & crop         & crop    & crop      & color-jitter 
		\\ \bottomrule
	\end{tabular}
\end{table}

\textbf{DAAC}. \cite{raileanu2021decoupling} For DAAC, we followed the implementation in the publicly released repository (\url{https://github.com/rraileanu/idaac}). DAAC was also implemented based on PPO and used two separate neural networks to perform the policy and value optimization. As reported in \cite{raileanu2021decoupling}, the number of epochs used during each update of the policy network was $1$, the number epochs used during each update of the value network was $9$, the value update frequency was $1$, the coefficient of the advantage loss was $0.25$, and the remaining hyperparameters were the same as Table~\ref{tb:procgen hyperparameters}.

\textbf{IDAAC}. \cite{raileanu2021decoupling} For IDAAC, we followed the implementation in the publicly released repository (\url{https://github.com/rraileanu/idaac}). IDAAC was a variant of DAAC that uses an auxiliary loss which constrains the policy representation to be invariant to the task instance. As reported in \cite{raileanu2021decoupling}, the number of epochs used during each update of the policy network was $1$, the number epochs used during each update of the value network was $9$, the value update frequency was $1$, the coefficient of instance-invariant (adversarial) loss was $0.001$, the coefficient of the advantage loss was $0.25$, and the remaining hyperparameters were the same as Table~\ref{tb:procgen hyperparameters}.

\textbf{DAAC+RE3}. \citep{seo2021state} For RE3, we followed the implementation in the publicly released repository (\url{https:
	//github.com/younggyoseo/RE3}). Here, the intrinsic reward is computed as $I_{t}=\log(\Vert\bm{e}_{t}-\tilde{\bm{e}}_{t}\Vert_{2}+1)$, where $\bm{e}_{t}=g(\bm{s}_{t})$ and $g$ is a random and fixed encoder. The total reward of time step $t$ is computed as $E_{t}+\beta_{t}\cdot I_{t}$, where $\beta_{t}=\beta_{0}(1-\kappa)^{t}$. Moreover, the average distance of $\bm{e}_t$ and its $k$-nearest neighbors was used to replace the single $k$ nearest neighbor to provide a less noisy state entropy estimate. As for hyperparameters related to exploration, we used $k=5$, $\beta_{0}=0.05$, and performed hyperparameter search over $\kappa\in\{0.00001, 0.000025\}$. Finally, the policy was updated using DAAC with general hyperparameters listed in Table~\ref{tb:procgen hyperparameters}.

\textbf{DAAC+RISE}. \cite{yuan2022renyi} For RISE, we followed the implementation in the publicly released repository (\url{https://github.com/yuanmingqi/rise}). Here, the intrinsic reward is computed as $I_{t}=(\Vert\bm{e}_{t}-\tilde{\bm{e}}_{t}\Vert_{2})^{1-\alpha}$, where $\bm{e}_{t}=g(\bm{s}_{t})$ and $g$ is a random and fixed encoder. The total reward of time step $t$ is computed as $E_{t}+\beta_{t}\cdot I_{t}$, where $\beta_{t}=\beta_{0}(1-\kappa)^{t}$. As for hyperparameters related to exploration, we used $k=5$, $\alpha=0.05$, $\beta_{0}=0.1$, and performed hyperparameter search over $\kappa\in\{0.00001, 0.000025\}$. Finally, the policy was updated using DAAC with general hyperparameters listed in Table~\ref{tb:procgen hyperparameters}.

\textbf{DAAC+RIDE}. \citep{raileanu2020ride} For RIDE, we followed the implementation in the publicly released repository (\url{https://github.com/facebookresearch/impact-driven-exploration}). In practice, we trained a single forward dynamics model $g$ to predict the encoded next-state $\psi(\bm{s}_{t+1})$ based on the current encoded state and action $(\psi(\bm{s}_{t}),\bm{a}_{t})$, whose loss function was $\Vert g(\psi(\bm{s}_{t}),\bm{a}_{t})-\psi(\bm{s}_{t+1})\Vert_{2}$. Then the intrinsic reward was computed as $I_{t}=\Vert\psi(\bm{s}_{t+1})-\psi(\bm{s}_{t})\Vert_{2}/\sqrt{N_{\rm ep}(\bm{s}_{t+1})}$,
where $N_{\rm ep}$ is the state visitation frequency during the current episode. To estimate the state visitation frequency of $\bm{s}_{t+1}$, we leveraged a pseudo-count method that approximates the frequency using the distance between $\psi(\bm{s}_{t})$ and its $k$-nearest neighbor within episode \citep{badia2020never}. Finally, the policy was updated using DAAC with general hyperparameters listed in Table~\ref{tb:procgen hyperparameters}.

\begin{table}[h]
	\centering
	\caption{General hyperparameters used to obtain the Procgen results.}
	\label{tb:procgen hyperparameters}
	\vskip 0.15in
	\begin{tabular}{ll}
		\toprule
		\textbf{Hyperparameter}             & \textbf{Value}         \\ \midrule
		Observation downsampling   & (84, 84)      \\
		Stacked frames             & No            \\
		Environment steps          & 25000000      \\
		Episode steps 			   & 265           \\
		Number of workers          & 1             \\
		Environments per worker    & 64            \\
		Optimizer                  & Adam          \\
		Learning rate              & 0.0005        \\
		GAE coefficient            & 0.95          \\
		Action entropy coefficient & 0.05          \\
		Value loss coefficient     & 0.5           \\
		Value clip range           & 0.2           \\
		Max gradient norm          & 0.5           \\
		Epochs per rollout         & 3             \\
		Mini-batches per epoch     & 8             \\
		Reward normalization       & Yes           \\
		LSTM                       & No            \\
		Gamma $\gamma$             & 0.99          \\ \bottomrule
	\end{tabular}
	\vskip -0.1in
\end{table}

\clearpage
\begin{table*}[h]
	\centering
	\caption{Procgen scores on train levels after training on 25M environment steps. (i) The mean and standard deviation are computed using 10 random seeds, and the highest average score is marked in color. (ii) The best data augmentation for each game is used when computing the results for DrAC. (iii) RE3 represents the combination of DAAC and RE3, RISE represents the combination of DAAC and RISE, and RIDE represents the combination of DAAC and RIDE. These three methods only use a fixed intrinsic reward function during the training, while AIRS automatically selects the most appropriate intrinsic reward in real-time. (iv) AIRS achieves the highest performance in 8 out of 16 games, especially in the \textit{StarPilot} game.}
	\label{tb:procgen results train}
	\vskip 0.15in
	\begin{tabular}{l|lllllll|l}
			\toprule
			Game      & PPO          & DrAC         & DAAC         & IDAAC        & RE3 & RISE & RIDE & AIRS (ours) \\ \midrule
			BigFish   & 10.2$\pm$2.0 & 10.2$\pm$1.4 & 20.0$\pm$2.9 & 19.7$\pm$1.1 & 19.4$\pm$1.9 & 17.0$\pm$1.3 & 19.1$\pm$2.2 & \textcolor[RGB]{247,127,0}{20.2$\pm$1.2} \\
			BossFight & 7.6$\pm$0.6  & 8.3$\pm$0.9  & 10.4$\pm$0.6 & 10.1$\pm$0.3 & 10.5$\pm$0.1 & 10.6$\pm$0.4 & 10.4$\pm$0.2 & \textcolor[RGB]{247,127,0}{10.7$\pm$0.4} \\
			CaveFlyer & 6.1$\pm$0.4  & \textcolor[RGB]{247,127,0}{6.6$\pm$1.2}  & 6.4$\pm$0.7  & 5.9$\pm$0.4  & 5.8$\pm$0.8  & 4.9$\pm$0.6  & 5.7$\pm$0.3  & 6.0$\pm$0.3  \\
			Chaser    & 4.7$\pm$0.5  & 5.5$\pm$1.1  & 5.1$\pm$0.3  & \textcolor[RGB]{247,127,0}{7.2$\pm$0.8}  & 5.2$\pm$0.7  & 4.1$\pm$1.1  & 5.6$\pm$0.5  & 5.4$\pm$0.4  \\
			Climber   & 8.2$\pm$0.6  & 8.8$\pm$0.8  & 9.4$\pm$0.7  & 9.3$\pm$0.4  & 9.2$\pm$0.5  & 9.6$\pm$0.4  & 9.3$\pm$0.2  & \textcolor[RGB]{247,127,0}{9.8$\pm$0.4}  \\
			CoinRun   & 8.8$\pm$0.2  & 9.4$\pm$0.2  & 9.9$\pm$0.2  & 9.9$\pm$0.2  & \textcolor[RGB]{247,127,0}{10.0$\pm$0.3}  & \textcolor[RGB]{247,127,0}{10.0$\pm$0.7}  & \textcolor[RGB]{247,127,0}{10.0$\pm$0.5}  & \textcolor[RGB]{247,127,0}{10.0$\pm$0.4}  \\
			Dodgeball & \textcolor[RGB]{247,127,0}{5.8$\pm$0.8}  & 4.5$\pm$0.5  & 4.6$\pm$0.5  & 4.6$\pm$0.4  & 4.4$\pm$0.7  & 3.2$\pm$0.2  & 4.4$\pm$0.3  & 5.4$\pm$0.4  \\
			FruitBot  & 27.4$\pm$0.4 & 29.8$\pm$0.7 & 29.6$\pm$1.0 & 29.0$\pm$0.4 & 30.0$\pm$1.1 & 28.9$\pm$1.5 & 29.5$\pm$1.8 & \textcolor[RGB]{247,127,0}{30.3$\pm$0.3} \\
			Heist     & 4.6$\pm$0.5  & \textcolor[RGB]{247,127,0}{8.0$\pm$0.4}  & 5.7$\pm$0.3  & 5.4$\pm$0.8  & 4.5$\pm$0.3  & 4.7$\pm$0.3  & 5.2$\pm$0.3  & 5.7$\pm$0.4  \\
			Jumper    & 8.6$\pm$0.3  & 8.6$\pm$0.3  & 8.5$\pm$0.5  & 8.5$\pm$0.3  & 8.3$\pm$0.4  & 7.6$\pm$0.3  & 8.4$\pm$0.7  & \textcolor[RGB]{247,127,0}{8.9$\pm$0.7}  \\
			Leaper    & 3.7$\pm$0.7  & 3.6$\pm$0.6  & 7.9$\pm$0.8  & \textcolor[RGB]{247,127,0}{8.4$\pm$0.9}  & 2.9$\pm$1.0  & 4.2$\pm$0.3  & 4.3$\pm$0.3  & 4.2$\pm$0.4  \\
			Maze      & 8.0$\pm$0.5  & \textcolor[RGB]{247,127,0}{8.6$\pm$0.3}  & 5.6$\pm$0.8  & 6.4$\pm$0.4  & 6.3$\pm$0.7  & 5.9$\pm$0.3  & 5.9$\pm$0.7  & 6.6$\pm$0.4  \\
			Miner     & 10.1$\pm$0.6 & 12.0$\pm$0.2 & 11.4$\pm$1.1 & 11.3$\pm$0.5 & 11.8$\pm$0.9 & 10.6$\pm$0.1 & 12.2$\pm$1.7 & \textcolor[RGB]{247,127,0}{12.4$\pm$0.2} \\
			Ninja     & 7.8$\pm$0.3  & 7.7$\pm$0.8  & 9.0$\pm$0.2  & 9.1$\pm$0.2  & 9.0$\pm$0.7  & 9.3$\pm$0.7  & \textcolor[RGB]{247,127,0}{9.6$\pm$0.9}  & 9.5$\pm$0.5  \\
			Plunder   & 6.3$\pm$0.4  & 6.2$\pm$1.1  & 23.0$\pm$1.7 & \textcolor[RGB]{247,127,0}{24.6$\pm$2.1} & 22.2$\pm$0.4 & 20.3$\pm$0.8 & 21.2$\pm$0.6 & 23.2$\pm$1.3 \\
			StarPilot & 30.3$\pm$1.9 & 31.2$\pm$2.9 & 38.3$\pm$2.0 & 34.3$\pm$2.0 & 40.0$\pm$3.0 & 36.0$\pm$1.3 & 36.0$\pm$2.1 & \textcolor[RGB]{247,127,0}{40.1$\pm$0.6} \\  \bottomrule
		\end{tabular}
	\vskip -0.1in
\end{table*}
\begin{table*}[h!]
	\centering
	\caption{Procgen scores on test levels after training on 25M environment steps. (i) The mean and standard deviation are computed using 10 random seeds, and the highest average score is marked in color. (ii) The best data augmentation for each game is used when computing the results for DrAC. (iii) RE3 represents the combination of DAAC and RE3, RISE represents the combination of DAAC and RISE, and RIDE represents the combination of DAAC and RIDE. These three methods only use a fixed intrinsic reward function during the training, while AIRS automatically selects the most appropriate intrinsic reward in real-time. (iv) AIRS achieves the highest performance in 9 out of 16 games, especially in the \textit{BigFish} game and \textit{StarPilot} game.}
	\label{tb:procgen results test}
	\vskip 0.15in
	\begin{tabular}{l|lllllll|l}
			\toprule
			Game      & PPO          & DrAC         & DAAC         & IDAAC        & RE3 & RISE & RIDE & AIRS (ours) \\ \midrule
			BigFish   & 4.2$\pm$0.9  & 8.2$\pm$1.1  & 17.0$\pm$3.6 & 17.9$\pm$0.7 & 18.0$\pm$1.3 & 15.2$\pm$4.9 & 16.5$\pm$3.5 & \textcolor[RGB]{247,127,0}{19.0$\pm$0.6} \\
			BossFight & 7.0$\pm$0.6  & 7.9$\pm$0.6  & 9.7$\pm$0.4  & 9.7$\pm$0.4  & 8.8$\pm$0.2  & 9.2$\pm$0.4  & 9.8$\pm$0.8  & \textcolor[RGB]{247,127,0}{9.9$\pm$0.4}  \\
			CaveFlyer & \textcolor[RGB]{247,127,0}{5.4$\pm$0.8}  & 5.3$\pm$0.5  & 4.7$\pm$0.4  & 5.1$\pm$0.3  & 4.7$\pm$0.6  & 3.8$\pm$0.5  & 4.7$\pm$0.7  & 4.8$\pm$0.4  \\
			Chaser    & 4.6$\pm$0.8  & 5.2$\pm$1.0  & 5.5$\pm$0.4  & 6.2$\pm$0.4  & \textcolor[RGB]{247,127,0}{6.2$\pm$0.7}  & 4.4$\pm$0.3  & 5.2$\pm$0.5  & \textcolor[RGB]{247,127,0}{6.2$\pm$0.4}  \\
			Climber   & 5.7$\pm$0.6  & 6.0$\pm$0.6  & 7.2$\pm$0.6  & \textcolor[RGB]{247,127,0}{7.8$\pm$0.3}  & 7.5$\pm$0.1  & 7.7$\pm$0.3  & 6.5$\pm$0.3  & 7.5$\pm$0.1  \\
			CoinRun   & 8.8$\pm$0.4  & 9.0$\pm$0.2  & 9.2$\pm$0.3  & 9.1$\pm$0.4  & 9.2$\pm$0.5  & \textcolor[RGB]{247,127,0}{9.9$\pm$0.6}  & \textcolor[RGB]{247,127,0}{9.9$\pm$0.5}  & \textcolor[RGB]{247,127,0}{9.9$\pm$0.3}  \\
			Dodgeball & 2.3$\pm$0.4  & 2.9$\pm$0.2  & 3.3$\pm$0.2  & 3.2$\pm$0.6  & 2.7$\pm$0.1  & 2.7$\pm$0.3  & 2.7$\pm$0.1  & \textcolor[RGB]{247,127,0}{3.4$\pm$1.0}  \\
			FruitBot  & 23.4$\pm$0.7 & 26.7$\pm$0.9 & 27.6$\pm$0.8 & 28.4$\pm$0.8 & 27.9$\pm$1.5 & 28.3$\pm$1.2 & 28.6$\pm$0.6 & \textcolor[RGB]{247,127,0}{30.0$\pm$0.4} \\
			Heist     & 2.8$\pm$0.7  & 3.6$\pm$0.5  & 3.5$\pm$0.5  & 3.0$\pm$0.3  & 3.4$\pm$0.1  & 3.3$\pm$0.6  & 3.4$\pm$0.3  & \textcolor[RGB]{247,127,0}{3.7$\pm$0.2}  \\
			Jumper    & 5.9$\pm$0.4  & 5.8$\pm$0.3  & 6.3$\pm$0.5  & 5.9$\pm$0.3  & \textcolor[RGB]{247,127,0}{6.7$\pm$0.3}  & 5.9$\pm$0.7  & 6.1$\pm$0.6  & 6.6$\pm$0.2  \\
			Leaper    & 2.9$\pm$0.4  & 3.7$\pm$0.6  & 7.0$\pm$0.5  & \textcolor[RGB]{247,127,0}{7.5$\pm$1.1}  & 3.6$\pm$0.3  & 4.0$\pm$0.3  & 3.6$\pm$0.7  & 4.2$\pm$0.7  \\
			Maze      & 5.6$\pm$0.7  & 5.1$\pm$0.5  & 5.5$\pm$1.1  & 5.6$\pm$0.6  & 5.7$\pm$1.0  & \textcolor[RGB]{247,127,0}{6.2$\pm$0.3}  & 5.4$\pm$0.3  & 5.9$\pm$0.5  \\
			Miner     & 7.6$\pm$0.7  & 9.1$\pm$0.4  & 8.5$\pm$0.8  & \textcolor[RGB]{247,127,0}{9.6$\pm$0.1}  & 8.8$\pm$1.3  & 9.1$\pm$0.6  & 8.7$\pm$1.0  & 9.0$\pm$0.5  \\
			Ninja     & 5.8$\pm$0.5  & 6.1$\pm$0.5  & 6.9$\pm$0.6  & 6.7$\pm$0.3  & 6.9$\pm$0.4  & 6.3$\pm$0.7  & 7.0$\pm$0.2  & \textcolor[RGB]{247,127,0}{7.5$\pm$0.3}  \\
			Plunder   & 5.9$\pm$0.6  & 9.4$\pm$0.4  & 20.6$\pm$1.5 & \textcolor[RGB]{247,127,0}{23.5$\pm$1.1} & 21.2$\pm$0.1 & 19.0$\pm$0.2 & 22.1$\pm$0.3 & 22.5$\pm$0.9 \\
			StarPilot & 24.7$\pm$1.3 & 27.0$\pm$3.4 & 36.8$\pm$1.7 & 35.9$\pm$3.4 & 36.7$\pm$2.0 & 35.6$\pm$3.0 & 37.1$\pm$2.4 & \textcolor[RGB]{247,127,0}{37.5$\pm$2.7} \\ \bottomrule
		\end{tabular}
	\vskip -0.1in
\end{table*}

\begin{figure*}[h]
	\vskip 0.2in
	\begin{center}
		\centerline{\includegraphics[width=\linewidth]{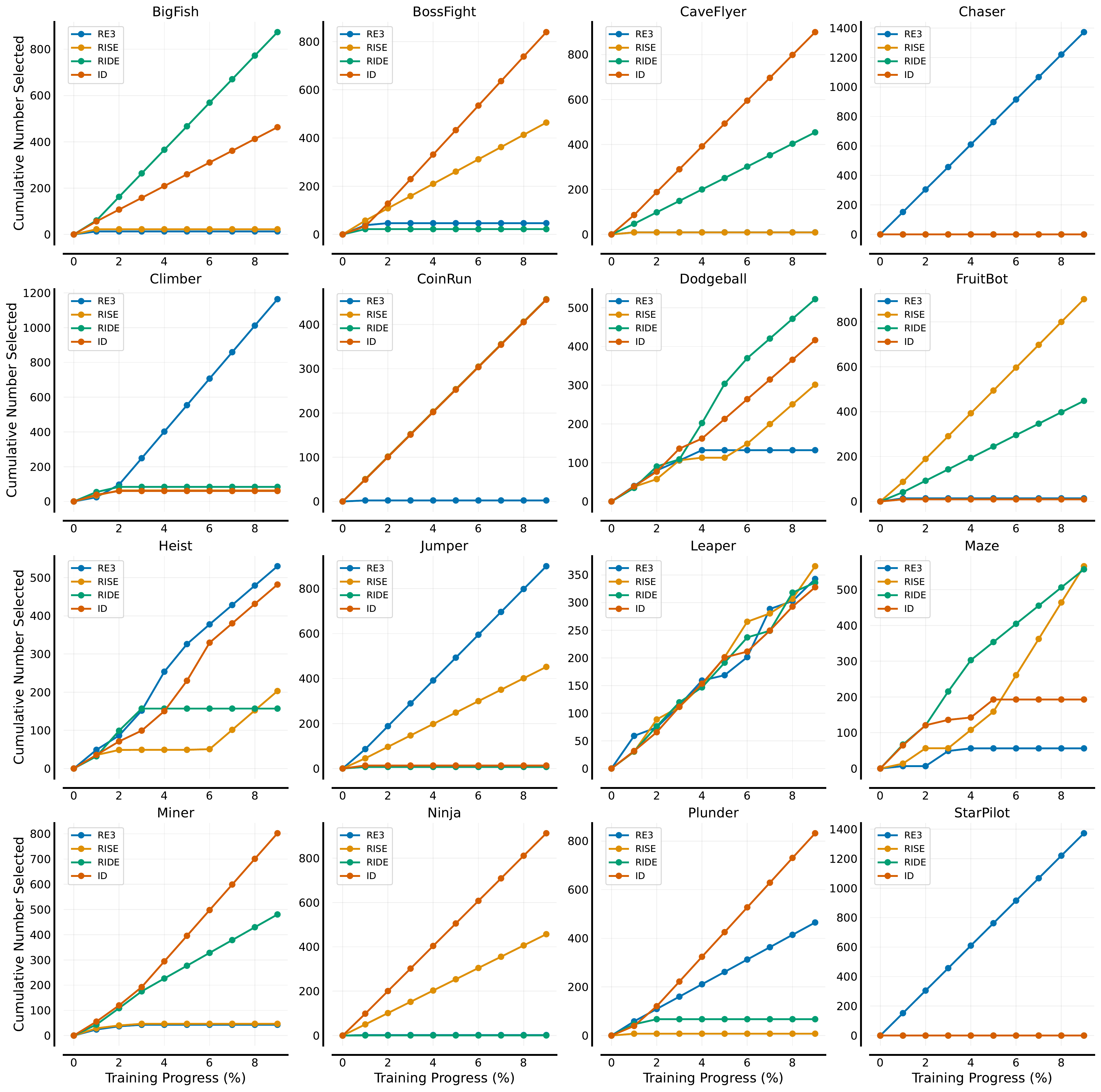}}
		\caption{The cumulative number of times AIRS selects each intrinsic reward function over the training progress, computed with ten random seeds.}
		\label{fig:procgen irs selection full}
	\end{center}
	\vskip -0.2in
\end{figure*}

\begin{figure*}[h]
	\vskip 0.2in
	\begin{center}
		\centerline{\includegraphics[width=\linewidth]{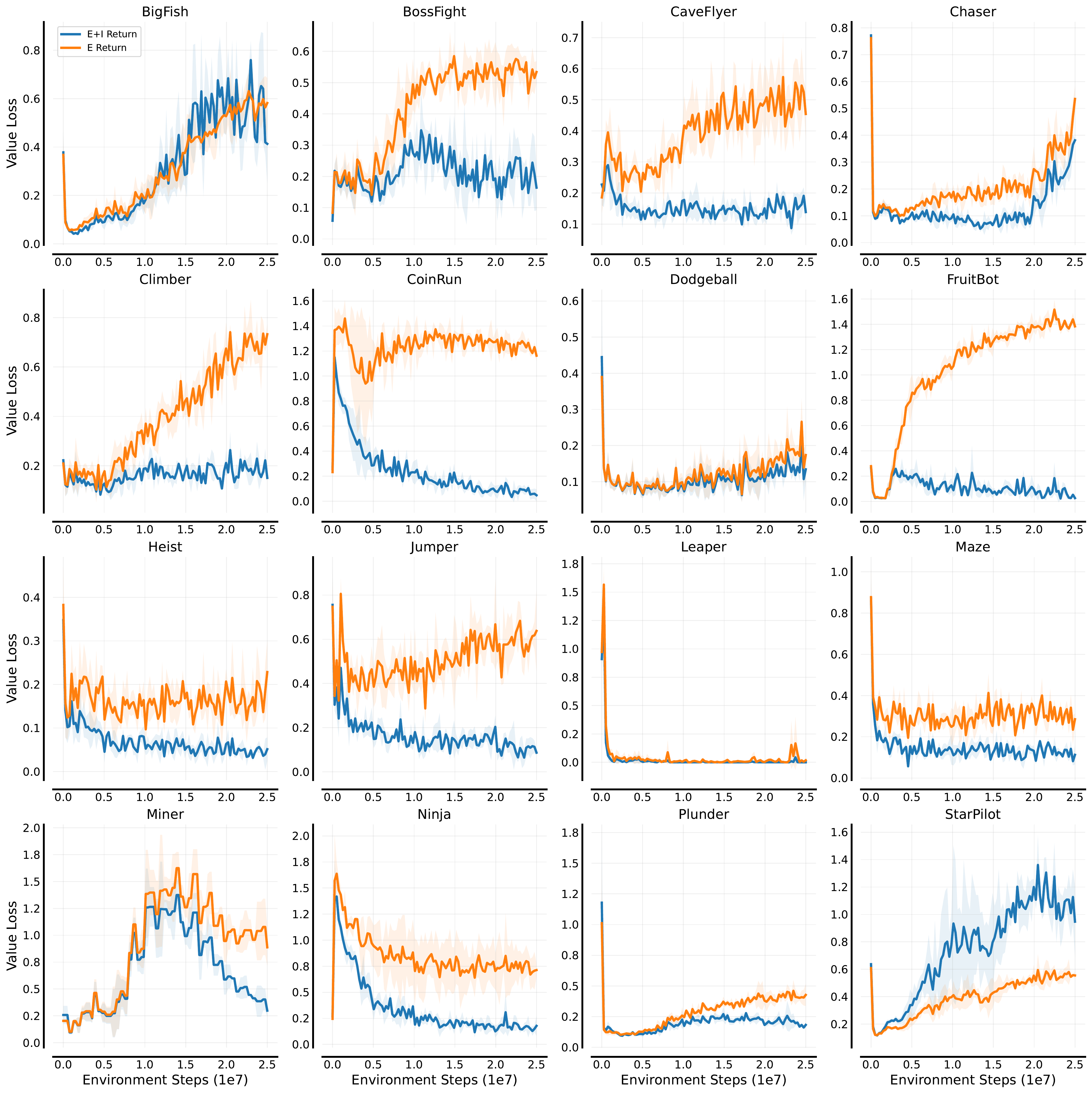}}
		\caption{The loss curves of value estimation. Here, "E Return" represents the estimated return evaluated by the extrinsic reward function. The solid line and shaded regions represent the mean and standard deviation over ten random seeds, respectively.}
		\label{fig:procgen value loss full}
	\end{center}
	\vskip -0.2in
\end{figure*}

\clearpage

\section{Details on DeepMind Control Suite Experiments}\label{appendix:dmc experiments}
\subsection{Environment Setting}
\begin{figure*}[th]
	\vskip 0.2in
	\begin{center}
		\centerline{\includegraphics[width=\linewidth]{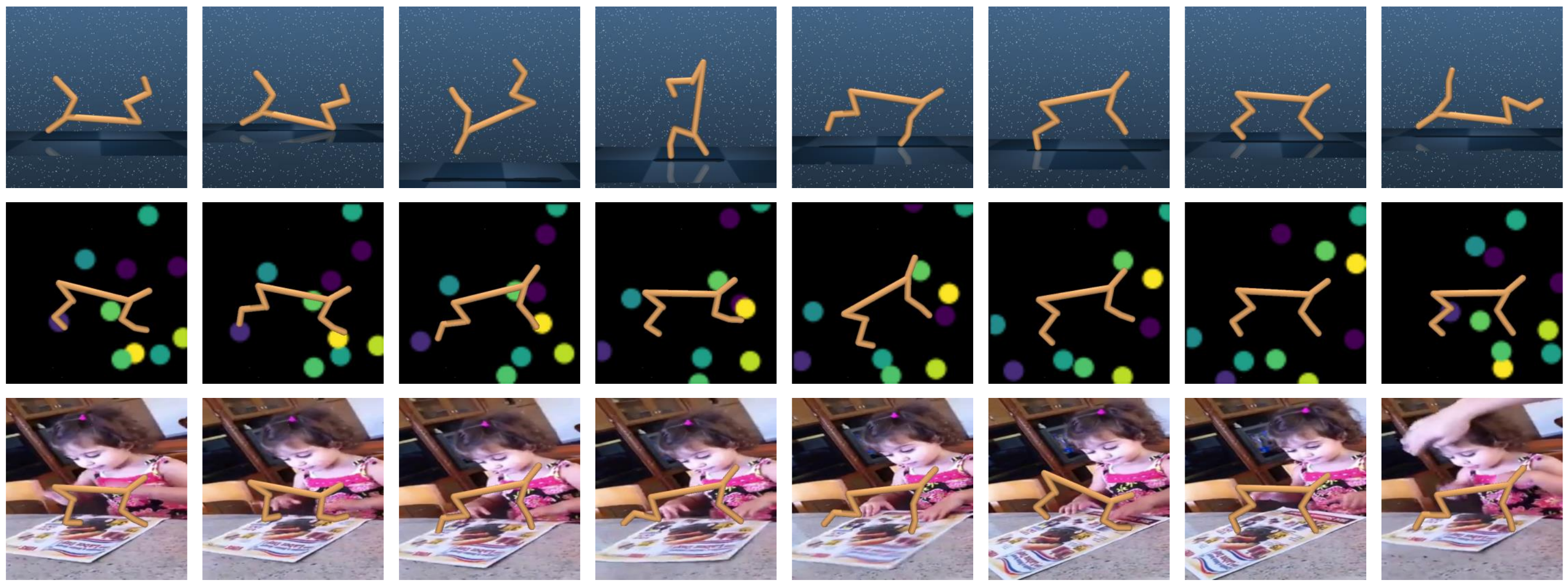}}
		\caption{Environment examples of DeepMind Control Suite. Top row: default backgrounds without any distractors. Middle row: synthetic distractor backgrounds with ideal gas videos. Bottom row: natural distractor backgrounds with Kinetics videos.}
		\label{fig:dmc examples}
	\end{center}
	\vskip -0.2in
\end{figure*}

In this section, we evaluated the performance of AIRS on four tasks from DeepMind Control Suite \cite{tassa2018deepmind}, namely \textit{Cartpole Balance}, \textit{Cheetah Run}, \textit{Finger Spin}, and \textit{Walker Walk}, respectively. To test the generalization ability of AIRS, we followed \cite{zhang2020learning} to replace the task background using synthetic distractors. Figure~\ref{fig:dmc examples} illustrates the derived observations with synthetic distractor backgrounds and default backgrounds, and the code for background generation can be found in (\url{https://github.com/facebookresearch/deep_bisim4control}). For each task, we stacked three consecutive frames as one observation, and these frames were further cropped to the size of (84, 84) to reduce the computational resource request.

\subsection{Experiment Setting}
We compared AIRS with PPO \cite{schulman2017proximal} and IDAAC \cite{raileanu2021decoupling}. For each task, the agents were trained for 1M frames and evaluated every 10K frames. Table~\ref{tb:dmc hyperparameters} illustrates the hyperparemeters derived by grid search. Any other hyperaparameters not mentioned here were set to the same values as the ones used for Procgen games in Table~\ref{tb:procgen hyperparameters}.

\begin{table}[h]
	\centering
	\caption{Hyperparameters used to obtain the DeepMind Control Suite results.}
	\label{tb:dmc hyperparameters}
	\vskip 0.15in
	\begin{tabular}{ll}
		\toprule
		\textbf{Hyperparameter}             & \textbf{Value}         \\ \midrule
		Observation downsampling   & (84, 84)      \\
		Stacked frames             & 3            \\
		Environment steps          & 1000000      \\
		Episode steps 			   & 2048           \\
		Number of workers          & 1             \\
		Environments per worker    & 1            \\
		Learning rate              & 0.001 \textit{Cartpole Balance}, \textit{Cheetah Run}, and \textit{Walker Walk}; 0.0001 \textit{Finger Spin}       \\
		GAE coefficient            & 0.99          \\
		Action entropy coefficient & 0.0001 \textit{Cheetah Run}; 0.0 \textit{Cartpole Balance} and \textit{Finger Spin}; 0.001 \textit{Walker Walk};       \\
		Epochs per rollout         & 0.0001 \textit{Cheetah Run}; 10 \textit{Cartpole Balance} and \textit{Finger Spin}; 5 \textit{Walker Walk}           \\
		Mini-batches per epoch     & 64 \textit{Cheetah Run}; 16 \textit{Cartpole Balance} and \textit{Finger Spin}; 32 \textit{Walker Walk}           \\ \bottomrule
	\end{tabular}
	\vskip -0.1in
\end{table}

\clearpage

\section{Intrinsic Reward Toolkit}\label{appendix:reward toolkit}

\subsection{Introduction}
Intrinsic rewards have been widely used to improve the exploration, and generalization ability of RL agents \cite{ryan2000intrinsic, csimcsek2006intrinsic, stadie2015incentivizing, bellemare2016unifying,  ostrovski2017count, burda2018exploration, savinov2018episodic, raileanu2020ride, yu2020intrinsic,yuan2022renyi,yuan2022rewarding}. However, there are great differences in the implementation of various intrinsic reward modules, which cannot provide efficient and reliable baselines. To facilitate our experiments and inspire subsequent research on intrinsic rewards, we developed a toolkit that provides high-quality implementations of various intrinsic reward modules based on PyTorch. This toolkit is designed to be highly modular and scalable. Each intrinsic reward module can be deployed in arbitrary algorithms in a plug-and-play manner. Table~\ref{tb:reward toolkit all} illustrates currently included implementations in the toolkit, and the code is available at \url{https://github.com/RLE-Foundation/rllte}.

\begin{table*}[h]
	\centering
	\caption{Included implementations of the intrinsic reward toolkit. (i) Here, $\bm{e}=\psi(\bm{s})$ is the learned representation of state $\bm{s}$, and $N_{\rm ep}$ is the episodic state visitation frequency. For ICM and RIDE, $\psi(\cdot)$ is learned by reconstructing the transition process. For RE3, RISE and REVD, $\psi(\cdot)$ is a random and fixed encoder. (iv) GIRM is a variant of ICM that utilizes variational auto-encoder \cite{kingma2013auto} to reconstruct the transition process and computes the intrinsic reward in a end-to-end manner. (iii) For RND, $\hat{f}$ is the target network that is fixed and randomly-initialized neural network, and $f$ is the predictor network that aims to approximate $\hat{f}$. (iv) The intrinsic reward produced by NGU is composed of episodic state novelty and life-long state novelty. Here, $\alpha_{t}$ is life-long curiosity factor computed following the RND method and $C$ is is a chosen maximum reward scaling. (v) For RE3 and RISE, $\tilde{\bm{e}}$ is the $k$-nearest neighbor of $\bm{s}$ in the encoding space. For REVD, $\tilde{\bm{e}}$ is the $k$-nearest neighbor of $\bm{s}$ within the current episode, and $\breve{\bm{e}}$ is the $k$-nearest neighbor of $\bm{s}$ within the former episode. (vi) PseudoCounts is an ablation of NGU in which the life-long module is deprecated.}
	\label{tb:reward toolkit all}
	\vskip 0.15in
	\begin{tabular}{lll}
		\toprule
		\textbf{Intrinsic reward module} & \textbf{Formulation} & \textbf{Remark} \\
		\midrule
		PseudoCounts \cite{badia2020never} & $I_{t}=1/\sqrt{N_{\rm ep}(\bm{s}_{t+1})}$ & Count-based exploration		 \\ \midrule
		ICM \cite{pathak2017curiosity} & $I_{t}=\Vert f(\bm{e}_{t},\bm{a}_{t})-\bm{e}_{t+1}\Vert_{2}^{2}$  & Curiosity-driven exploration \\ \midrule
		RND \cite{burda2018exploration} & $I_{t}=\Vert \hat{f}(\bm{s}_{t+1})-f(\bm{s}_{t+1})\Vert_{2}^{2}$ & Count-based exploration \\ \midrule
		GIRM \cite{yu2020intrinsic} & $I_{t}=\Vert \hat{\bm{s}}_{t}-\bm{s}_{t+1}\Vert_{2}^{2}$ & Curiosity-driven exploration	 \\ \midrule
		NGU \cite{badia2020never} & $I_{t}=\min\{\max\{\alpha_{t}\}, C\}/\sqrt{N_{\rm ep}(\bm{s}_{t+1})}$ & Memory-based exploration		 \\ \midrule
		RIDE \cite{raileanu2020ride} & 
		$I_{t}=\Vert\bm{e}_{t+1}-\bm{e}_{t}\Vert_{2}/\sqrt{N_{\rm ep}(\bm{s}_{t+1})}$
		& Significant state changes\\ \midrule
		RE3 \cite{seo2021state}    & $I_{t}=\frac{1}{k}\sum_{i=1}^{k}\log(\Vert\bm{e}_{t}-\tilde{\bm{e}}_{t}^{i}\Vert_{2}+1)$  & Shannon entropy maximization\\ \midrule
		RISE \cite{yuan2022renyi}  &  
		$I_{t}=\frac{1}{k}\sum_{i=1}^{k}(\Vert\bm{e}_{t}-\tilde{\bm{e}}_{t}^{i}\Vert_{2})^{1-\alpha}$ & R\'enyi entropy maximization \\ \midrule
		REVD \cite{yuan2022renyi}  &  
		$I_{t}=\frac{1}{k}\sum_{i=1}^{k}(\Vert\bm{e}_{t}-\breve{\bm{e}}_{t}^{i}\Vert_{2}/\Vert\bm{e}_{t}-\tilde{\bm{e}}_{t}^{i}\Vert_{2})^{1-\alpha}$ & R\'enyi divergence maximization \\ 
		\bottomrule
	\end{tabular}
	\vskip -0.1in
\end{table*}

\subsection{Usage Example}
Due to the large differences in the calculation of different intrinsic reward methods, the toolkit has the following rules:
\begin{itemize}
	\item The environments are assumed to be \textcolor[RGB]{247,127,0}{gym-like} and \textcolor[RGB]{247,127,0}{vectorized};
	\item Each intrinsic reward module has a \textcolor[RGB]{247,127,0}{compute\_irs} function with a mandatory argument \textcolor[RGB]{247,127,0}{rollouts}. It is a Python dict like \cite{harris2020array}:
	\begin{table*}[h]
		\centering
		\vskip 0.15in
		\begin{tabular}{lll}
			\toprule
			\textbf{Key} & \textbf{Data shape}  & \textbf{Data type}   \\ \midrule
			observations & (n\_steps, n\_envs, *obs\_shape)   & PyTorch Tensor \\
			actions      & (n\_steps, n\_envs, *action\_shape) & PyTorch Tensor \\
			rewards      & (n\_steps, n\_envs)             & PyTorch Tensor \\ 
			next observations & (n\_steps, n\_envs, *obs\_shape)   & PyTorch Tensor \\ \bottomrule
		\end{tabular}
		\vskip -0.1in
	\end{table*}
\end{itemize}

Take RE3 for instance, it computes the intrinsic reward for each state $\bm{s}$ based on the Euclidean distance between the state $\bm{s}$ and its $k$-nearest neighbor within a mini-batch. Thus it suffices to provide \textcolor[RGB]{247,127,0}{observations} data to compute the reward. The following code provides a usage example of RE3:

\begin{figure}[h!]
	\vskip 0.2in
	\begin{center}
		\centerline{\includegraphics[width=0.9\linewidth]{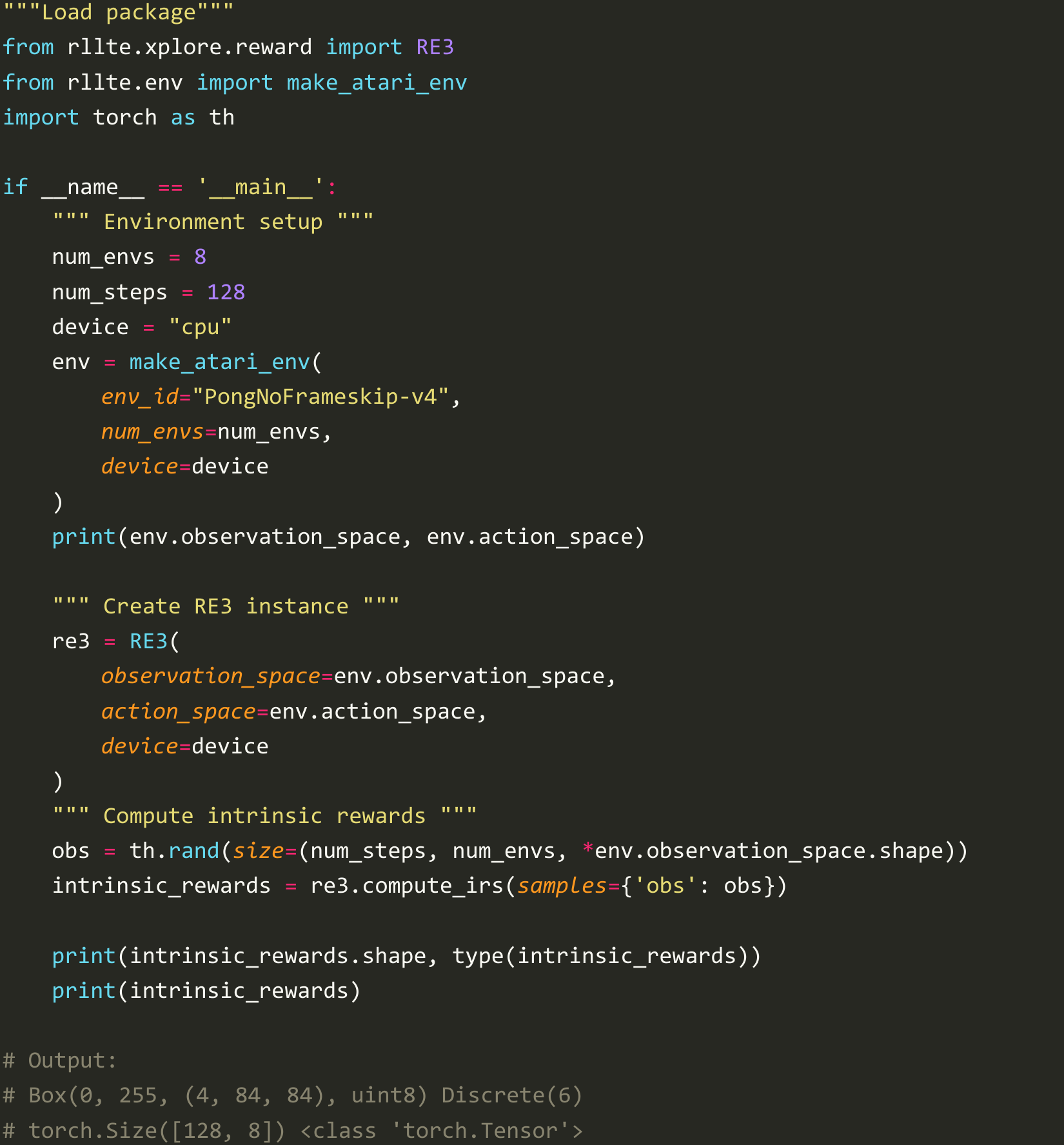}}
		\caption{Code example of the intrinsic reward toolkit.}
		\label{fig:code example}
	\end{center}
	\vskip -0.2in
\end{figure}

We are testing this toolkit on various tasks (e.g., OpenAI Gym and DeepMind Control Suite) and will establish a complete test dataset to provide convenient baselines. More consequent results can be found at \url{https://hub.rllte.dev/}. We will also continue to follow up the latest research on intrinsic reward-driven exploration and provide reliable implementations.

\end{document}